\documentclass[review]{elsarticle}
\makeatletter
\def\ps@pprintTitle{%
 \let\@oddhead\@empty
 \let\@evenhead\@empty
 \def\@oddfoot{}%
 \let\@evenfoot\@oddfoot}
\makeatother
\usepackage{lineno,hyperref}
\usepackage{amssymb}
\usepackage{tikz}
\usepackage{tikz-qtree}
\usepackage[utf8]{inputenc}
\usepackage[top=1.6in, bottom=1.6in, left=1.3in, right=1.3in]{geometry}
\usepackage{tcolorbox}
\usepackage{xcolor}
\usepackage{algorithm}
\usepackage{amsmath}
\usepackage[noend]{algpseudocode}
\usepackage{adjustbox}
\usepackage{setspace}
\newtcolorbox[auto counter]{example}[1][]{
  fonttitle=\scshape,
  title={Example \thetcbcounter},
  #1
}

\usepackage{color, colortbl}
\definecolor{Gray}{gray}{0.9}

\usepackage{pgfplots}
\pgfplotsset{width=10cm,compat=1.15}
\usepgfplotslibrary{statistics}

\usepackage{listings}

\journal{Artificial Intelligence}
\biboptions{authoryear}

\begin{document}

\begin{frontmatter}

\title{Optimizing Readability Using Genetic Algorithms}

\author{Jorge Martinez-Gil}
\address{Software Competence Center Hagenberg GmbH \\ Softwarepark 32a, 4232 Hagenberg, Austria \\ \url{jorge.martinez-gil@scch.at}}

\begin{abstract}
This research presents ORUGA, a method that tries to automatically optimize the readability of any text in English. The core idea behind the method is that certain factors affect the readability of a text, some of which are quantifiable (number of words, syllables, presence or absence of adverbs, and so on). The nature of these factors allows us to implement a genetic learning strategy to replace some existing words with their most suitable synonyms to facilitate optimization. In addition, this research seeks to preserve both the original text's content and form through multi-objective optimization techniques. In this way, neither the text's syntactic structure nor the semantic content of the original message is significantly distorted. An exhaustive study on a substantial number and diversity of texts confirms that our method was able to optimize the degree of readability in all cases without significantly altering their form or meaning. The source code of this approach is available at \url{https://github.com/jorge-martinez-gil/oruga}
\end{abstract}


\end{frontmatter}

\section{Introduction}
Readability is a measure that tells us how easy it is to read a text. It corresponds to the level of literacy that is expected from the readers in the target audience. In this way, readability is considered one of the most critical factors that facilitate the user experience when consuming information. It is crucial because it is key to establishing a trusting relationship between information producers and consumers. It must be considered that some factors, such as complexity, legibility, or typography, contribute to making a text readable. However, not all factors are quantifiable and cannot be optimized by automatic techniques. In this paper, we focus solely and exclusively on factors of a quantifiable nature, which always revolve around basic or advanced statistics associated with the text to be optimized.

Therefore, text readability refers to how simple it is to read and comprehend a given text, depending on its unique characteristics. These characteristics are usually measurable through metrics like the number of syllables in a sentence. The diversity of words used to create a readability score can be used to gauge this measure \citep{collins2014computational}. Therefore, ORUGA is intended to facilitate the reader's ability to understand a text by optimizing its readability. 

In the context of this work, a clear distinction between readability and quality should be made. Quality emphasizes essential elements like grammar, spelling, and voice. However, the goal of producing textual information as plainly as possible and better matching it with its audience is what is meant when a text is said to be readable. In this way, text readability partially overlaps with the notion of text difficulty, which is also an essential aspect of human language, and has an impact on the daily lives of most people who consume written information.

Our research is based on the premise that readability can be measured by considering metrics related to the text and then using a specific mathematical formula to calculate it. Because different readability scores are calculated using different mathematical formulas, it is possible to design a strategy that replaces many of the terms in the text with synonyms using a global optimization scheme. Such optimization is about maximizing or minimizing the result yielded by such mathematical formulas at the user's convenience. In practice, such a strategy can be implemented through a genetic algorithm, as shown throughout this work. 

Therefore, our research offers a viewpoint from the field of computer science, concentrating on the fundamental text representations and metrics utilized by readability assessment methods. In this way, the most significant contributions of this work to state-of-the-art are the following:

\begin{itemize}
	\item We present, for the first time, a technique that can automatically optimize the readability of any text, i.e., we can minimize or maximize the degree of readability of a text automatically without substantive changes.
	\item We study which are the best sources of synonyms currently available for text readability optimization. Specifically, we have studied Wordnet, word2vec, and web scraping and established a classification around optimizing up to ten texts of different natures.
	\item We present an additional method based on multi-objective optimization, whose mission is to ensure that the minimum number of words needed in the original text is replaced in such a way that the structure of the text is not impacted.
	\item Last but not least, we studied several strategies that allow us to measure (and therefore optimize) the semantic distance between the original text and the generated text. In this way, the impact on the original text's meaning can be controlled.
\end{itemize}

This research work is structured as follows: Section 2 shows state-of-the-art methods and tools for improving text readability. Section 3 presents the technical details of our proposal; these technical details are based on the design of a genetic cutting strategy that allows us to explore a vast search space while consuming a reasonable amount of resources. Section 4 explains how to minimize the impact on the form of the original text using a multi-objective optimization technique. Section 5 shows how to preserve the essence of the original text by ensuring that the distance between the original text and the text obtained is kept as small as possible. It is necessary to remark that sections 3, 4, and 5 present information about the design of different experiments and the raw results obtained, and their subsequent analysis. Finally, we conclude with the main lessons that can be learned from this research.

\section{State-of-the-art}
Let us begin with a formal definition of readability; for instance,\citep{chall1995readability} defines readability as the "\textit{total number of elements in a given text that affect a reader's success}." This reader's success is a measure of how well a text that is read at an optimal speed can be understood. At the same time,  \citep{mc1969smog} defines readability as "\textit{the level at which certain people find reading material convincing and understandable}." Beyond these definitions, we are particularly interested in the quantifiable aspects of readability, i.e., what can be objectively measured. Otherwise, it would not be easy to proceed to its optimization using a computer. Let us see what the literature says about these quantifiable aspects.

\subsection{Text readability in the scientific literature}
Readability metrics usually use simple features to calculate the degree of readability of a text. Some commonly used features are the number of sentences or words, the ratio of unique words, the total number of syllables, the proportion of unique words in the text, the number of digits, the number of words with many syllables, etc. Although, at first, it may seem that these metrics are simplistic, they are very commonly used for two important reasons: they are much cheaper and faster to use than the alternatives consisting of human surveys, and according to experts, they usually give exceptionally reliable results that are in line with reality.

The goal of improving readability is to increase the chances that readers can understand the thoughts and ideas reflected in the text. So that misunderstanding is minimized, information processing is facilitated without requiring much effort and energy consumption. With this goal in mind, many sources can be found that advise how to improve a text's readability. However, these are manually compiled protocols that a human operator must translate into reality by modifying the text manually. For example, for a given metric, it is better to shorten sentences; for another, it is better to replace complex words with simpler ones, etc. 

This is precisely where our contribution to the state-of-the-art lies. Optimizing texts automatically using a metric as a target means we do not have to concern about taking any manual action leading to altering the text. The genetic algorithm will find a way to proceed automatically. 

\subsection{Why is text readability important?}
There are several contexts and population groups for which readability is critical. Especially when it is necessary to convey a written message to an audience. For example,

\begin{itemize}
	\item Teachers need to be sure of the readability of a text before deciding whether it is appropriate for their students. This is particularly important in language learning. With the method presented here, the text can be optimized for a certain niche of learners.

	\item In the world of advertising, readability allows for building a trust relationship between advertisers and potential consumers. Advertising goods or services using texts with high readability is usually not a good idea since the message might not reach an essential part of the population. This is even more important in the search engine optimization sector, as many search engines use readability metrics as a ranking factor when responding to user searches. 

	\item Readability is also relevant for professionals who work on websites \citep{key-Pantula}, news \citep{key-Qin}, or even educational materials \citep{key-Ante}. In some countries, there is even a legal requirement that government agencies provide textual information with certain readability levels to reach the entire population.
\end{itemize}

\subsection{Readability metrics}
There are several metrics to quantify how readable a text is \citep{meade1991readability}. Most metrics have been designed for the English language \citep{key-Maqsood}, although works also explore readability in other languages \citep{madrazo2020cross}. Without being exhaustive, we can mention, in chronological order, some of the metrics that enjoy or have enjoyed more significant popularity when dealing with the English language.

Text readability depends not only on the characteristics of the text but also on the educational background of the individuals interested in understanding the text. We will see this reflected below when measuring readability using formulas. We will see how readability metrics take a text as input and calculate a numerical score that usually corresponds to the level of education required to understand the text.

\subsubsection{Dale-Chall readability}
The Dale-Chall readability formula \citep{dale1948formula} requires a list of 3,000 words that fourth-grade U.S. students could reliably understand, as shown in Equation \ref{eq:dcrf}. 

\begin{equation}
	DCRF = 0.1579 \left(\frac{difficult \ words}{total \ words} \times 100\right) + 0.0496 \left(\frac{total \ words}{total \ sentences} \right)
	\label{eq:dcrf}
\end{equation}

\subsubsection{SMOG readability}
The SMOG readability level \citep{mc1969smog} can be assessed through a formula originally used for checking health messages, as shown in Equation \ref{eq:smog}. It corresponds to the years of education necessary to understand the text.

\begin{equation}
	SMOG = 1.0430 \sqrt{number \ of \ polysyllables\times{\frac{30}{number \ of \ sentences}} } + 3.1291
	\label{eq:smog}
\end{equation}

\subsubsection{ARI readability}
The ARI assesses the U.S. grade level required to read a text \citep{senter1967automated}. In some ways, it is similar to other formulas. Its difference is that rather than counting syllables, it counts characters: the more characters, the more complex the word. It also counts sentences as shown in Equation \ref{eq:ari}. This sets it apart from some other formulas.

\begin{equation}
	ARI=4.71\left(\frac{total \ Characters}{total \ Words}\right)+0.5\left(\frac{total \  Words}{total \  Sentences}\right)-21.43
  \label{eq:ari}
\end{equation}

\subsubsection{Flesch Kincaid readability}
Flesch Kincaid's readability score, as shown in Equation \ref{eq:fkgl}, is a metric based on grade levels that is used commonly in the insurance industry. Grade levels made it much easier for people to understand \citep{kincaid1975derivation}. A Flesch Kincaid Grade Level (FKGL) between 8 and 10 means that the text should be accessible to the public. FKGL remains the most widely-used formula today.

\begin{equation}
	FKGL =	206.835 - 1.015 \left( \frac{total \ words}{total \ sentences} \right) - 84.6 \left( \frac{total \ syllables}{total \ words} \right)
	\label{eq:fkgl}
\end{equation}

Over the past decade, several natural language processing (NLP) techniques have been proposed to determine the readability of a text. Thus, as opposed to the classical approach of using formulas that measure a limited set of text features, these new variants have attempted to measure the difficulty of understanding sentences and words and even the complexity of the syntax \citep{martinc2021supervised}. Even some techniques based on predictors of readability, such as cohesion and coherence, have received considerable attention. However, so far, these approaches have yet to be able to predict the readability of a text better than the classical techniques discussed here \citep{todirascu2016cohesive}.

\subsection{Semantic Similarity}
The field of semantic similarity measurement \citep{key-martinez-mlwa} is one of the most active in several different research communities (information retrieval, database integration, natural language processing, and so on) \citep{key-Rus}. This is due to its significant implications on many available frameworks, methods, and tools \citep{key-Navigli} both in industry and academia. The literature around this topic has skyrocketed in the last few years \citep{key-chandrasekaran}. However, most research works focus on determining the similarity between words \citep{key-sematch}, phrases, or documents \citep{key-martinez-eswa,key-martinez-eswa2}. 

In this way, rarely the likelihood of effectively throwing out the most similar words to a given one has been discussed. To this effect, there are synonym libraries that have been manually compiled and some word embedding techniques that do the job well. The most prominent solutions in this direction are WordNet \citep{key-Pedersen}, word2vec \citep{key-Mikolov}, or BERT \citep{key-Bert}. However, it must be taken into account that the resource requirements for techniques based on the computation of word embeddings are very high \citep{key-martinez-jfis}. 

\subsection{Contribution over the state-of-the-art}
Determining text readability based on a formal analysis of the structures and words used has been a recurring theme in the literature over the last few years. As a result, many metrics have been proposed to measure text readability. A common denominator of all these metrics is that a high score usually means the text is difficult to understand. In other words, a higher degree of study is needed to understand it. For this reason, many communication professionals often use tools to help them discern whether the text fits a given audience. However, none of these tools can do the professional's job automatically, which is why our contribution is essential. 

To the best of our knowledge, this is the first time anyone has proposed automatically improving the readability of text without significantly altering the content or the text's form. In addition, we put this innovative method through its paces using a wide range of texts that varied in subject matter and level of readability. Furthermore, we provide the source code of the first implementation for anyone interested in experimenting with or improving the method.

\section{Part I: Design and Implementation of a Functional Solution}
\label{sec:partI}
In the following, we explain our proposal for text readability optimization using a method based on genetic algorithms. In this section, we outline the technical preliminaries, discuss the implementation, show an illustrative example of how our approach works in practice, and conduct an experimental study using real data and use cases that can help us get an idea of the performance of this approach in practice.

\subsection{Technical Preliminaries}
Our hypothesis is that we can find a set of synonyms to replace some words in the original text so that the value returned by the readability formula can be optimized. For example, if the readability formula rewards or penalizes long words, we have to find synonyms of less length. In reality, we only have to concern about understanding which formula best represents readability in our specific scenario and indicate it as a fitness function. The genetic algorithm will \textit{understand} how to proceed. Thus, we are faced with a classical optimization problem.

We intend to act only on the vocabulary since it is one of the most critical parts of that language. It is widely assumed that vocabulary is the essential part of a language because, without vocabulary, it is impossible to compose any message \citep{wilkins1972linguistics}. We do not act on proper nouns, prepositions, or other stop words to avoid distorting the original message.

While this problem can be solved by a brute-force search over the range of the words of a given text $w_0, w_1,...,w_n$, the GA method scales very well when dealing with large texts. In this case, a brute-force search would be prohibitively expensive. We could act on other aspects, such as the structure, but then we would risk distorting the original text's essence again. 

Our idea is to find the combination of words (which will be encoded in the form of an individual) that optimizes the desired objective. The choice of the fitness function is effortless and has the advantage that the solution automatically identifies what kind of words lead to the optimization.

In this work, we work with three main sources of synonyms:

\begin{itemize}
	\item \textit{WordNet} \citep{miller1995wordnet}, which is a thesaurus that has been manually compiled. It is probably one of the most widely used dictionaries in information systems and will give us several alternatives to substitute each candidate word. No surprises are to be expected in this library, except perhaps the substitution of words with a synonym that has a sense far from the original. In any case, we have a solution to this problem.
	\item \textit{wordvec} \citep{key-Mikolov}, which is an approach that calculates the vectors associated with each word according to a textual corpus of relevance. Once each vector has been computed, a computation process can be performed by which the $N$ vectors most similar to a given one are identified. This way, synonyms of relevance are obtained for the word in question. Note that this process is computationally expensive.
	\item \textit{Web scraping} \citep{mitchell2018web}, which consists of obtaining the synonyms from some websites, usually specialized, so that we can shuffle several alternatives per candidate word. This method obtains many high-quality word candidates, but it should be used responsibly because it can cause problems on the server side if many hundreds or thousands of requests are made.  Therefore, the method is fine for experimentation, but it would be unreasonable to exploit it.
\end{itemize}

\subsection{Implementation}
The implementation of this novel approach is based on a classical optimization scheme using genetic algorithms. Algorithm \ref{alg:the_alg} briefly explains in pseudo-code how the whole process is performed by adapting the classical structure of the genetic algorithm \citep{key-forrest} where different operators capable of implementing selection, cross-over and mutation processes are considered in order to evolve a given population towards the desired objective.

\begin{algorithm}
\caption{Optimizing Readability Using Genetic Algorithm}
\label{alg:the_alg}
\begin{algorithmic}[1]
\Procedure{ORUGA}{}
\State \textit{population $\gets$ generationRandomIndividual (population)}
\State \textit{calculateReadabilityScore (population)}
\State \textbf{while} \textit{(stop condition not reached)} \textbf{do}
\State \ \ \ \ \textit{parents $\gets$ selectionOfIndividuals (population)}
\State \ \ \ \ \textit{offspring $\gets$ Crossover (parents)}
\State \ \ \ \ \textit{offspring $\gets$ Mutation (offspring)}
\State \ \ \ \ \textit{offspring $\gets$ calculateReadabilityScore (offspring)}
\State \ \ \ \ \textit{population $\gets$ updatePopulation (offspring)}
\State \textbf{endwhile}
\State optimizedText $\gets$ \textit{optimizedIndividual (population)}
\State optimizedText $\gets$ \textit{correctErrorsIfNecessary (optimizedText)}
\State \textbf{return} \textit{optimizedText}
\EndProcedure
\end{algorithmic}
\end{algorithm}

Please note that before starting the evolution process, a pre-processing of the text must be done in order to identify the candidate words to be replaced by a synonym. We propose that all words are candidates except: proper names, stop words, and prepositions. The reason for this is that they are often words for which it is really difficult to find a synonym.

In relation to the genetic algorithm itself, the parametric details of the solution will be discussed below, but it is possible to see how we implement it in the form of a classical evolutionary process. This means that a population of individuals is selected randomly, and their readability score is calculated. We then proceed with an iterative process of selection, crossover, and mutation; the best individuals are passed from generation to generation until one of the stopping criteria is met: the highest possible has been reached (unlikely), or the number of iterations has been exhausted (very likely). 

Finally, at the end of the evolutionary process, we correct the text in case grammatical errors are produced by substituting a synonym that does not fit the tense and the form of the sentence in which it is framed. This way, we obtain a corrected readability score, which may vary slightly from the one derived automatically. In return, we ensure that the results are usable, or at least close to being usable.

\subsection{Illustrative examples}
Example \ref{ex:A} shows us how ORUGA works with written material about science extracted from Wikipedia. In fact, our aim is to observe how ORUGA behaves when trying to minimize the FKGL readability score using synonyms from the library WordNet. Let us remember that FKGL (or one of its variants) is probably the most widely used metric and its optimization brings advantages in several fields. Please note that the text to be treated can be of any length, but to facilitate the presentation to the reader, we have chosen (and will choose throughout this paper) one that contains only several sentences.

\begin{example}[label=ex:A,colback=black!5!white,colframe=black!40!black,title=Example 1. Science]
\begin{tcolorbox}[colback=black!10!white,colframe=black!70!black,title=Original text \textbf{Source:} Wikipedia,colbacktitle=black!50!white]
``The sea moderates the climate and has important roles in the water cycle, carbon cycle, and nitrogen cycle. Humans harnessing and studying the sea have been recorded since ancient times, and evidenced well into prehistory, while its modern scientific study is called oceanography. The most abundant solid dissolved in seawater is sodium chloride. The water also contains salts of magnesium, calcium, potassium, and mercury, amongst many other elements, some in minute concentrations. Salinity varies widely, being lower near the surface and the mouths of large rivers and higher in the depths of the ocean; however, the relative proportions of dissolved salts vary little across the oceans.'' \textcolor{blue}{\textbf{FKGL score: 14.06}}.
\end{tcolorbox}
\begin{tcolorbox}[colback=black!10!white,colframe=black!70!black,title=ORUGA - minimizing the FKGL score - library Wordnet,colbacktitle=black!50!white]
The sea moderates the climate and has important \textcolor{red}{part} in the \textcolor{red}{H2O} cycle, \textcolor{red}{C} cycle, and \textcolor{red}{N} cycle. Humans harnessing and studying the sea have been \textcolor{red}{taped} since ancient times, and \textcolor{red}{attested} well into prehistory, while its modern scientific study is called oceanography. The most abundant solid \textcolor{red}{fade out} in \textcolor{red}{brine} is \textcolor{red}{Na} chloride. The \textcolor{red}{H2O too} contains salts of magnesium, calcium, potassium, and mercury, amongst many other elements, some in \textcolor{red}{min} concentrations. Salinity \textcolor{red}{changes} widely, being \textcolor{red}{got down} near the surface and the mouths of large rivers and higher in the \textcolor{red}{depth} of the ocean; however, the relative proportions of \textcolor{red}{fade out} salts \textcolor{red}{change} little across the oceans. \textcolor{blue}{\textbf{FKGL score: 11.85}}
\end{tcolorbox}
\end{example}

As can be seen, ORUGA can optimize the textual input by first automatically identifying which words can be replaced by a synonym, and then undertaking a process of searching for synonyms that improve the results of the metric to be optimized. In this way, the impact on the initial message is minimal, although it is true that some synonyms can slightly distort the meaning of the text, and therefore final supervision by the user is required. However, there is no need to concern because this problem will be addressed later in the paper.

Let us look now at Example \ref{ex:B}, which is a written text about the history of Austria that has been also extracted from Wikipedia, and we would like to to minimize the FKGL readability score using synonyms automatically obtained by web scraping.

\begin{example}[label=ex:B,colback=black!5!white,colframe=black!40!black,title=Example 2. History]
\begin{tcolorbox}[colback=black!10!white,colframe=black!70!black,title=Original text \textbf{Source:} Wikipedia,colbacktitle=black!50!white]
``Austria emerged from the remnants of the Eastern and Hungarian March at the end of the first millennium. Originally a margraviate of Bavaria, it developed into a duchy of the Holy Roman Empire in 1156 and was later made an archduchy in 1453. In the 16th century, Vienna began serving as the empire administrative capital and Austria thus became the heartland of the Habsburg monarchy. After the dissolution of the Holy Roman Empire in 1806, Austria established its own empire, which became a great power and the dominant member of the German Confederation. The defeat in the Austro-Prussian War of 1866 led to the end of the Confederation and paved the way for the establishment of Austria-Hungary a year later.'' \textcolor{blue}{\textbf{FKGL score: 13.20}}
\end{tcolorbox}
\begin{tcolorbox}[colback=black!10!white,colframe=black!70!black,title=ORUGA - minimizing the FKGL score - web scraping,colbacktitle=black!50!white]
Austria \textcolor{red}{looms} from the \textcolor{red}{debris} of the Eastern and Hungarian March at the end of the first millennium. Originally a \textcolor{red}{margravate} of Bavaria, it \textcolor{red}{matured within} a duchy of the Holy Roman Empire in 1156 and was \textcolor{red}{next} made an \textcolor{red}{arch duchy} in 1453. In the 16th century, Vienna \textcolor{red}{lead plate} as the \textcolor{red}{command departmental central} and Austria thus \textcolor{red}{come} the heartland of the Habsburg monarchy. After the \textcolor{red}{divorce} of the Holy Roman Empire in 1806, Austria \textcolor{red}{settled} its own empire, that \textcolor{red}{come} a great power and the dominant \textcolor{red}{branch} of the German Confederation. The defeat in the Austro-Prussian War of 1866 led to the end of the Confederation and \textcolor{red}{brick} the way for the \textcolor{red}{founding} of Austria-Hungary a year later. \textcolor{blue}{\textbf{FKGL score: 11.35}}
\end{tcolorbox}
\end{example}

Once again, we can see how the genetic algorithm has acted intelligently to decrease the text readability score and therefore make the text accessible to more people. It is clear that the suitability of some words may be subject to debate, but the first objective of this research, i.e., to optimize the readability score, has been achieved. As we mentioned before, we will be concerned to outline a final product later in this paper.

Finally, let us look at the opposite case, i.e., let us see if we can make a text more difficult to read without losing its essence. In Example \ref{ex:C}, we have a text about sports also extracted from Wikipedia, and we do not want to make the text accessible to as many people as possible, but on the contrary, we want to increase the level of readability. We will now try a different metric, e.g., ARI.

\begin{example}[label=ex:C,colback=black!5!white,colframe=black!40!black,title=Example 3. Sports]
\begin{tcolorbox}[colback=black!10!white,colframe=black!70!black,title=Original text \textbf{Source:} Wikipedia,colbacktitle=black!50!white]
``Real Madrid Club de Futbol, meaning Royal Madrid Football Club, commonly referred to as Real Madrid, is a Spanish professional football club based in Madrid. Founded in 1902 as Madrid Football Club, the club has traditionally worn a white home kit since its inception. The honorific title real is Spanish for Royal and was bestowed to the club by King Alfonso XIII in 1920 together with the royal crown in the emblem. Real Madrid have played their home matches in the Santiago Bernabeu Stadium in downtown Madrid since 1947. Unlike most European sporting entities, Real Madrid members (socios) have owned and operated the club throughout its history.'' \textcolor{blue}{\textbf{ARI score: 12.69}}.
\end{tcolorbox}
\begin{tcolorbox}[colback=black!10!white,colframe=black!70!black,title=ORUGA - maximizing the ARI score - web scraping,colbacktitle=black!50!white]
Real Madrid Club de Futbol, \textcolor{red}{connotation} Royal Madrid Football Club, \textcolor{red}{frequently} referred to as Real Madrid, is a Spanish professional football \textcolor{red}{business established} in Madrid. Founded in 1902 as Madrid Football Club, the \textcolor{red}{business} has \textcolor{red}{consistently timeworn} an \textcolor{red}{alabaster} home kit since its inception. The \textcolor{red}{appellation} title real is Spanish for Royal and was \textcolor{red}{entrusted} to the \textcolor{red}{business} by King Alfonso XIII in 1920 together \textcolor{red}{alongside} the \textcolor{red}{aristocratic culmination} in the emblem. Real Madrid have played their \textcolor{red}{familiar} matches in the Santiago Bernabéu Stadium in downtown Madrid \textcolor{red}{afterward} 1947. Unlike most European sporting entities, Real Madrid \textcolor{red}{assemblage} (socios) have owned and \textcolor{red}{negotiated} the \textcolor{red}{business} throughout its history. \textcolor{blue}{\textbf{ARI score: 15.53}}
\end{tcolorbox}
\end{example}

As can be seen, after processing the fragment related to \textit{Real Madrid} extracted from Wikipedia, the genetic algorithm selects synonyms that are much longer and more complicated to read to increase the ARI score. Although a use case consisting of making the text less accessible to people is hard to imagine, it may find application in some specific niches of learning, education, etc.

\subsection{Experimental study}
In this section, we explain the details of an empirical study that we have carried out to know the feasibility of our new method. To do so, we first established the conditions of the experiments by setting up an experimental setup. Secondly, we have run and obtained the raw results. Furthermore, thirdly, and lastly, we have proceeded with the analysis of the data obtained.

\subsubsection{Experimental setup}
Adjusting the parameters of the genetic algorithm differs from the focus of this work. Therefore, we have chosen a classical parameter setting, which has been shown to work quite well. Standard tuning of the parameters, e.g., through a grid search, is a possible line of future work. In the meantime, the parameters we have used for our experiments are the following: 

\begin{itemize}
	\item Population size \{10, 15, 20\}: \textbf{20}
	\item Number of parents mating \{10, 15, 20\}: \textbf{10}
	\item Number of genes: one per candidate word to be substituted by a synonym
	\item Fitness function: the user can choose among Equations \ref{eq:dcrf}, \ref{eq:smog}, \ref{eq:ari}, \ref{eq:fkgl}
	\item Stop condition: \{100, 200, 300\}: \textbf{300} generations
\end{itemize}

To test whether our approach can optimize text readability, we semi-randomly selected ten texts extracted from Wikipedia. These texts are classified into several categories engineering, geography, history, science, and sports. Some metrics, such as SMOG, require at least 30 sentences to apply their formula. For cases where we do not reach 30 sentences, we will duplicate the text until we reach that number of sentences. It should also be noted that depending on which readability category each use case represents. For example, magenta represents the highest difficulty (FKGL between 15 and 18), equivalent to an academic paper. Red represents medium-high difficulty (FKGL between 12 and 15). The black color represents a medium difficulty (FKGL between 9 and 12). Furthermore, the blue color represents a low-medium difficulty. It is estimated that up to 80\% of the native English-speaking population could understand text with an FKGL between 6 and 9, which is what the blue value represents.

Furthermore, every experiment was run on a standard computer with 32 GB of primary memory and an Intel Core i7-8700 processor running at 3.20 GHz on Microsoft Windows 10 64-bit. Most of the functionality has been implemented using the library PyGAD\footnote{https://pypi.org/project/pygad/}, an open-source Python library for building genetic algorithms.

\subsubsection{Experiments}
The first experiment is the one that is the most useful in practice. It consists of trying to minimize the readability of the ten proposed texts so that the processed text can reach the largest possible audience. To do that, we want to use readability metrics which estimate the readability of a text based on simple aspects such as syllable and word counts. We use the FKGL score since this metric, or some of its variants, has been used for decades on traditional texts, and it is still one of the most common and widely used traditional readability measures.

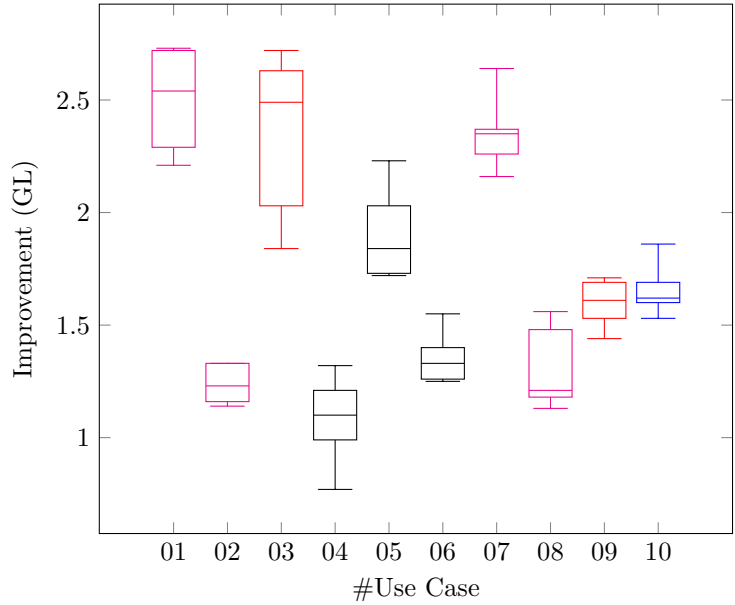
\begin{figure}
	\centering
\begin{tikzpicture}
\begin{axis}[
boxplot/draw direction=y,
xlabel={\#Use Case},
ylabel={Improvement (GL)},
xtick={1,2,3,4,5,6,7,8,9,10},
xticklabels={01, 02, 03, 04, 05, 06, 07, 08, 09, 10},
]
\addplot[color=magenta] [boxplot prepared={draw position=1,
lower whisker=2.21, lower quartile=2.29,
median=2.54, upper quartile=2.72,
upper whisker=2.73},
] coordinates {};
\addplot[color=magenta]  [boxplot prepared={draw position=2,
lower whisker=1.14, lower quartile=1.16,
median=1.23, upper quartile=1.33,
upper whisker=1.33},
] coordinates {};
\addplot[color=red]  [boxplot prepared={draw position=3,
lower whisker=1.84, lower quartile=2.03,
median=2.49, upper quartile=2.63,
upper whisker=2.72},
] coordinates {};
\addplot[color=black]  [boxplot prepared={draw position=4,
lower whisker=0.77, lower quartile=0.99,
median=1.10, upper quartile=1.21,
upper whisker=1.32},
] coordinates {};
\addplot[color=black]  [boxplot prepared={draw position=5,
lower whisker=1.72, lower quartile=1.73,
median=1.84, upper quartile=2.03,
upper whisker=2.23},
] coordinates {};
\addplot[color=black] [boxplot prepared={draw position=6,
lower whisker=1.25, lower quartile=1.26,
median=1.33, upper quartile=1.40,
upper whisker=1.55},
] coordinates {};
\addplot[color=magenta]  [boxplot prepared={draw position=7,
lower whisker=2.16, lower quartile=2.26,
median=2.35, upper quartile=2.37,
upper whisker=2.64},
] coordinates {};
\addplot[color=magenta]  [boxplot prepared={draw position=8,
lower whisker=1.13, lower quartile=1.18,
median=1.21, upper quartile=1.48,
upper whisker=1.56},
] coordinates {};
\addplot[color=red]  [boxplot prepared={draw position=9,
lower whisker=1.44, lower quartile=1.53,
median=1.61, upper quartile=1.69,
upper whisker=1.71},
] coordinates {};
\addplot[color=blue]  [boxplot prepared={draw position=10,
lower whisker=1.53, lower quartile=1.60,
median=1.62, upper quartile=1.69,
upper whisker=1.86},
] coordinates {};
\end{axis}
\end{tikzpicture}
\caption{Results for the \textbf{minimization} of the \textbf{FKGL} score using \textbf{WordNet}}
\label{fig:f1}
\end{figure}

Fig \ref{fig:f1} shows us the results obtained when reducing the FKGL for the texts under consideration using the library WordNet. The results shown are the summary of ten independent runs per use case. Since we are dealing with cold start methods with random values, and there is even a component of randomness in the mutations, we will almost always get a different result, so it is essential to show the results in this way.

Positive results have been obtained in all 100 experiments performed (ten runs for ten use cases). In addition, minimum improvements, average improvements, and even maximum improvements have been achieved in the range between 0.77 and 2.73 points on the FKGL scale. 

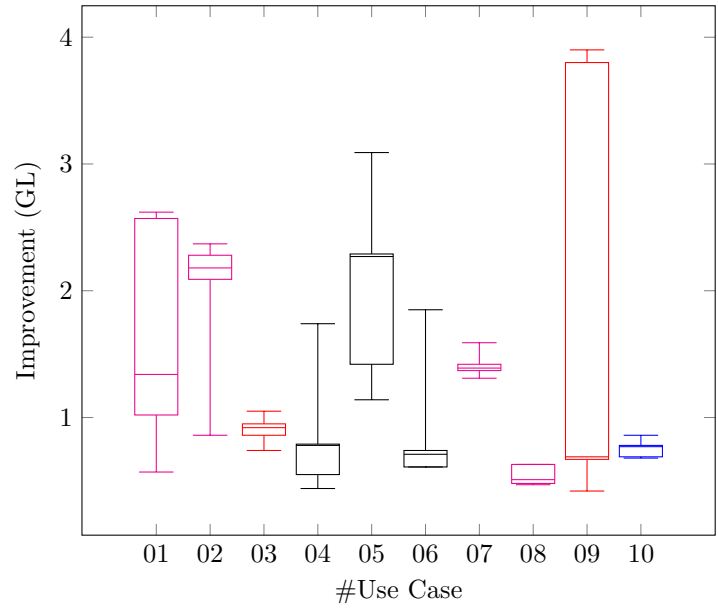
\begin{figure}
	\centering
\begin{tikzpicture}
\begin{axis}[
boxplot/draw direction=y,
xlabel={\#Use Case},
ylabel={Improvement (GL)},
xtick={1,2,3,4,5,6,7,8,9,10},
xticklabels={01, 02, 03, 04, 05, 06, 07, 08, 09, 10},
]
\addplot[color=magenta] [boxplot prepared={draw position=1,
lower whisker=0.57, lower quartile=1.02,
median=1.34, upper quartile=2.57,
upper whisker=2.62},
] coordinates {};
\addplot[color=magenta]  [boxplot prepared={draw position=2,
lower whisker=0.86, lower quartile=2.09,
median=2.18, upper quartile=2.28,
upper whisker=2.37},
] coordinates {};
\addplot[color=red]  [boxplot prepared={draw position=3,
lower whisker=0.74, lower quartile=0.86,
median=0.92, upper quartile=0.95,
upper whisker=1.05},
] coordinates {};
\addplot[color=black]  [boxplot prepared={draw position=4,
lower whisker=0.44, lower quartile=0.55,
median=0.78, upper quartile=0.79,
upper whisker=1.74},
] coordinates {};
\addplot[color=black]  [boxplot prepared={draw position=5,
lower whisker=1.14, lower quartile=1.42,
median=2.27, upper quartile=2.29,
upper whisker=3.09},
] coordinates {};
\addplot[color=black] [boxplot prepared={draw position=6,
lower whisker=0.61, lower quartile=0.61,
median=0.71, upper quartile=0.74,
upper whisker=1.85},
] coordinates {};
\addplot[color=magenta]  [boxplot prepared={draw position=7,
lower whisker=1.31, lower quartile=1.37,
median=1.39, upper quartile=1.42,
upper whisker=1.59},
] coordinates {};
\addplot[color=magenta]  [boxplot prepared={draw position=8,
lower whisker=0.47, lower quartile=0.48,
median=0.51, upper quartile=0.63,
upper whisker=0.63},
] coordinates {};
\addplot[color=red]  [boxplot prepared={draw position=9,
lower whisker=0.42, lower quartile=0.67,
median=0.69, upper quartile=3.8,
upper whisker=3.9},
] coordinates {};
\addplot[color=blue]  [boxplot prepared={draw position=10,
lower whisker=0.68, lower quartile=0.69,
median=0.77, upper quartile=0.78,
upper whisker=0.86},
] coordinates {};
\end{axis}
\end{tikzpicture}
\caption{Results for the \textbf{minimization} of the \textbf{FKGL} score using \textbf{word2vec}}
\label{fig:f2}
\end{figure}

Fig \ref{fig:f2} shows us the results obtained using the synonym calculation method using word2vec. As it is possible to see, the variance of the results is enormous, which means that using this library will make the results not very predictable. At the same time, as in the previous case, all 100 experiments were able to obtain readability improvements that ranged between 0.42 and 3.90-grade levels.

\begin{figure}
	\centering
\begin{tikzpicture}
\begin{axis}[
boxplot/draw direction=y,
xlabel={\#Use Case},
ylabel={Improvement (GL)},
xtick={1,2,3,4,5,6,7,8,9,10},
xticklabels={01, 02, 03, 04, 05, 06, 07, 08, 09, 10},
]
\addplot[color=magenta] [boxplot prepared={draw position=1,
lower whisker=1.75, lower quartile=1.78,
median=1.89, upper quartile=2.09,
upper whisker=2.1},
] coordinates {};
\addplot[color=magenta]  [boxplot prepared={draw position=2,
lower whisker=1.64, lower quartile=1.65,
median=1.74, upper quartile=1.84,
upper whisker=1.89},
] coordinates {};
\addplot[color=red]  [boxplot prepared={draw position=3,
lower whisker=1.46, lower quartile=1.85,
median=1.87, upper quartile=1.93,
upper whisker=1.94},
] coordinates {};
\addplot[color=black]  [boxplot prepared={draw position=4,
lower whisker=1.02, lower quartile=1.08,
median=1.09, upper quartile=1.10,
upper whisker=1.23},
] coordinates {};
\addplot[color=black]  [boxplot prepared={draw position=5,
lower whisker=2.06, lower quartile=2.17,
median=2.33, upper quartile=2.4,
upper whisker=2.43},
] coordinates {};
\addplot[color=black] [boxplot prepared={draw position=6,
lower whisker=1.42, lower quartile=1.49,
median=1.54, upper quartile=1.58,
upper whisker=1.59},
] coordinates {};
\addplot[color=magenta]  [boxplot prepared={draw position=7,
lower whisker=2.56, lower quartile=3.23,
median=3.31, upper quartile=4.02,
upper whisker=4.02},
] coordinates {};
\addplot[color=magenta]  [boxplot prepared={draw position=8,
lower whisker=1.35, lower quartile=1.4,
median=1.49, upper quartile=1.71,
upper whisker=1.72},
] coordinates {};
\addplot[color=red]  [boxplot prepared={draw position=9,
lower whisker=1.69, lower quartile=1.7,
median=1.77, upper quartile=1.82,
upper whisker=1.85},
] coordinates {};
\addplot[color=blue]  [boxplot prepared={draw position=10,
lower whisker=1.11, lower quartile=1.12,
median=1.17, upper quartile=1.19,
upper whisker=1.29},
] coordinates {};
\end{axis}
\end{tikzpicture}
\caption{Results for the \textbf{minimization} of the \textbf{FKGL} score using \textbf{Web Scraping}}
\label{fig:f3}
\end{figure}
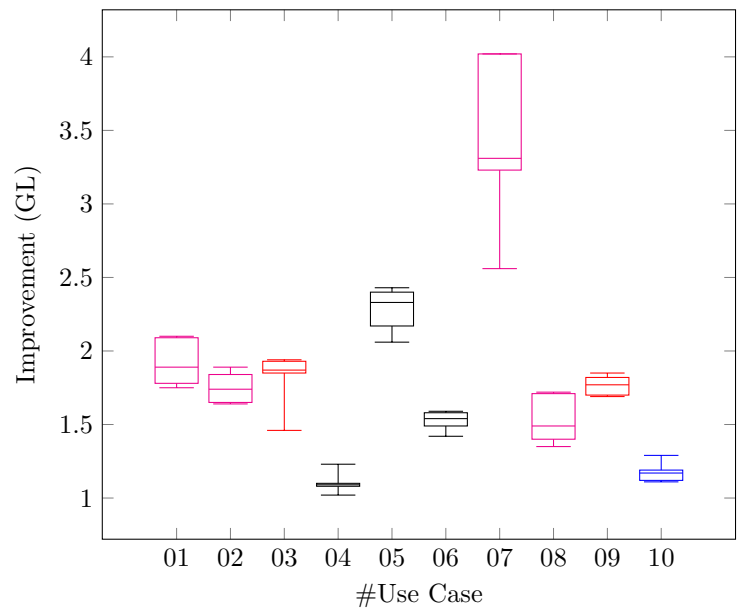

Fig \ref{fig:f3} shows us the results we have obtained by searching for synonyms with web scraping techniques. The results this time have a smaller variance than in the previous case. In addition, we have again obtained favorable results in all 100 experiments performed. The optimization achieved ranges between 1.02 and 4.20-grade levels.

Table \ref{tab:t0} shows the summary results of the 300 experiments performed (100 for each synonym library). For WordNet, a median optimization of 1.63-grade levels is expected, very similar to that achieved by web scraping. The median optimization using word2vec is the worst of the three libraries considered. Please note that we are not judging here the quality of the replacements, but the values to optimize the input text independently of the quality of the final result.

\begin{table}
	\centering
	\scalebox{0.95}{
\begin{tabular}{|c|c|c|c|}
\hline
Library      & Minimum & Median & Maximum \\
\hline
WordNet      & 0.77    &   1.63     & 2.73    \\
word2vec     & 0.42    &   0.96     & 3.90     \\
Web scraping & 1.02    &   1.61     & 4.20    \\
\hline     
\end{tabular}}
	\caption{Summary of the results obtained using the different libraries of synonyms}
	\label{tab:t0}
\end{table}

Table \ref{tab:t1} shows a summary of all the results we have obtained. The texts have been randomly obtained from Wikipedia. The range of values shown indicates the minimum value that could be reached and the maximum value (for minimization and maximization problems respectively).

We analyze the four formulas that we consider most representative, but the analysis of other formulas would not be a problem. When working with SMOG, we must be careful because the text to be analyzed requires at least 30 sentences. We have duplicated the text for these experiments so that these 30 sentences can be used.

\begin{table}
		\centering
		\scalebox{0.95}{
    \begin{tabular}{|r|c|c|c|c|}
        \hline
        \textbf{Use case} & \textbf{DCRF} & \textbf{SMOG}  & \textbf{ARI} & \textbf{FKGL} \\ \hline
				01/science & 1.11 $\pm$ 0.17	&	2.03 $\pm$ 0.13 & 2.90 $\pm$ 0.27 	& 2.53 $\pm$ 0.15 \\
				02/history & 	0.29 $\pm$ 0.12 	&	0.71 $\pm$ 0.08  & 0.84 $\pm$ 0.05	& 0.90 $\pm$ 0.04   \\
				03/geography &  0.65 $\pm$ 0.07	&	 0.91 $\pm$ 0.04 & 1.17 $\pm$ 0.12	&  1.33 $\pm$ 0.04  \\
				04/sports & 0.78 $\pm$ 0.14	&	 0.42 $\pm$ 0.12 & 0.83 $\pm$ 0.05 	&  0.93 $\pm$ 0.10 \\
				05/engineering & 0.59 $\pm$	0.06 & 0.60 $\pm$ 0.10	 & 0.66 $\pm$ 0.10	&  0.75 $\pm$ 0.09 \\
				06/geography & 0.52 $\pm$ 0.15	&	 0.70 $\pm$ 0.10 & 0.77 $\pm$	0.19 &  0.77 $\pm$ 0.18 \\
				07/engineering  & 1.03 $\pm$ 0.15	&	0.51 $\pm$ 0.03 & 1.12 $\pm$ 0.21	&  1.32 $\pm$ 0.01 \\
				08/history & 0.77 $\pm$ 0.09	&	0.27 $\pm$ 0.06 & 0.44 $\pm$ 0.00	&  0.73 $\pm$ 0.17 \\
				09/sports & 0.46 $\pm$ 0.06	&	1.06 $\pm$ 0.28 & 0.99 $\pm$ 0.12	&  0.81 $\pm$ 0.10 \\
				10/history & 0.74 $\pm$ 0.04 &	0.65 $\pm$ 0.03 & 0.77 $\pm$ 0.13	&  0.93 $\pm$ 0.04 \\
        \hline
    \end{tabular}}
	\caption{Summary of the results obtained for the experiments performed w.r.t. \textbf{minimization}. The numerical values represent absolute improvements after using ORUGA}
	\label{tab:t1}
\end{table}

Table \ref{tab:t2} shows a summary of all the results we have obtained when maximizing the readability scores. As can be seen by comparing with the table above, it is much easier to increase the level of readability than to decrease it, i.e., it is easier to make a text difficult to read than the other way around.

\begin{table}
		\centering
		\scalebox{0.95}{
    \begin{tabular}{|r|c|c|c|c|}
        \hline
        \textbf{Use case} & \textbf{DCRF} & \textbf{SMOG}  & \textbf{ARI} & \textbf{FKGL} \\ \hline
				01/science &   1.98 $\pm$ 0.16 	& 1.30 $\pm$ 0.06  & 1.80 $\pm$ 0.08	& 2.09 $\pm$ 0.14  \\
				02/history & 1.85 $\pm$ 0.25	& 1.99	$\pm$ 0.09 & 3.88 $\pm$ 0.28 	& 3.02 $\pm$ 0.08  \\
				03/geography & 	2.38 $\pm$ 0.12 	&	1.23 $\pm$ 0.08  & 4.44 $\pm$ 0.13	& 4.00 $\pm$ 0.44  \\
				04/sports & 2.09 $\pm$ 0.40 	& 2.79 $\pm$ 0.07 & 4.63 $\pm$ 0.40 	& 4.26 $\pm$ 0.04   \\
				05/engineering & 2.10 $\pm$ 0.43 	& 2.16 $\pm$ 0.11  & 4.55 $\pm$ 0.10	& 4.23 $\pm$ 0.35  \\
				06/geography & 	1.53 $\pm$ 0.21 	& 1.81 $\pm$ 0.22  & 2.96 $\pm$ 0.04	& 2.90 $\pm$ 0.41   \\
				07/engineering & 1.49 $\pm$ 0.38	&	1.71 $\pm$ 0.04  & 3.87 $\pm$ 0.46	& 3.34 $\pm$ 0.16   \\
				08/history & 	1.43 $\pm$ 0.18 	&	3.09 $\pm$ 0.13  & 4.11 $\pm$ 0.63	& 3.70 $\pm$ 0.21   \\
				09/sports & 1.60 $\pm$ 0.21 	&	2.29 $\pm$ 0.16  & 3.54 $\pm$ 0.28	& 2.56 $\pm$ 0.04   \\
				10/history & 2.06 $\pm$ 0.08	&	2.64 $\pm$ 0.04  & 3.99 $\pm$ 0.25	& 3.27 $\pm$ 0.52   \\
        \hline
    \end{tabular}}
	\caption{Summary of the results obtained for the experiments performed w.r.t. \textbf{maximization}. The numerical values represent absolute improvements after using ORUGA}
	\label{tab:t2}
\end{table}

\subsubsection{Discussion}
We have seen how it is possible to build a solution that optimizes the readability of texts of different natures. Moreover, such optimization is done respecting the content and the form of such texts, trying to minimize the impact of word replacements by synonyms that better fit the readability criteria of the different formulas we have studied. We have seen how optimization can occur in two ways. First, it can be done in such a way as to reduce the readability score, which will allow more people to understand the text perfectly. This is undoubtedly the most practical option in practice. Furthermore, secondly, it can be done to increase the readability score, allowing a smaller number of people to understand the processed text unambiguously. This option has less practical utility than can be discerned at first glance.

Based on the research we have carried out, possible improvements can be made. For example, a multi-objective optimization algorithm could simultaneously optimize the readability score by affecting as few words as possible. In this way, the impact of our method on the original text would be even more negligible. A solution front could allow the human operator to decide the trade-off between the score modification and the replaced words.

Finally, and as a limitation of our method, it can be observed that sometimes the processed text contains minor grammatical errors. This is because we have yet to use techniques that allow, for example, to choose the appropriate verb tense for the context. Here there are two alternatives. On the one hand, we can implement better methods for correcting grammatical errors to obtain a corrected readability score; on the other hand, we can give the user the option to edit or choose the word that best fits each moment.

\section{Part II: Minimizing the impact on the form of the original text}
\label{sec:partII}
While in Section \ref{sec:partI}, we have seen that it is possible to design a functional solution to optimize the text readability, we have also seen that the approach can be intrusive at times. That is, the replacement of many words by synonyms can lead to a distortion of the original message. For this reason, in this section, we focus on minimizing the impact of ORUGA on the original text by replacing as few words as possible. To do so, we will build a solution based on multi-objective optimization (MOO) that allows us to optimize readability and simultaneously minimize the number of replacements. This section comprises the technical preliminaries, the implementation, some illustrative examples, and an empirical study to test several texts of different natures.

\subsection{Technical Preliminaries}
We have already seen how in the field of text readability, there has been a great effort to build straightforward formulas that can be understood by people and have a good correlation to how easily a text can be read from a human perspective. Now we go one step further to obtain a higher quality result. MOO is a strategy in which two or more objectives are simultaneously optimized. This is the situation we find ourselves in, given that we want: on the one hand, to improve the readability of a text, and on the other hand, we want to reduce as much as possible the number of words that need to be replaced to improve readability, and thus to minimize the impact that our approach has on the original text form. 

Furthermore, MOO is useful when decisions must be made despite potential trade-offs between more than one orthogonal objective. Again, this is our situation because our goals of maximizing readability while simultaneously replacing the fewest possible words require us to pursue two completely different goals. In situations like this, different solutions can simultaneously fulfill all objectives. As a result, all optimal solutions ought to be regarded as equivalent merit without any external evaluation from a human operator \citep{key-martinez-kbs}. More formally, we can model a MOO problem as expressed in Equation \ref{eq:moo}.

\begin{equation}
	\begin{gathered}
	\min \left(s_1(\vec{x}), s_2(\vec{x}), \ldots, s_n(\vec{x}) \right)  \\
   subject \ to \ \vec{x}\in X
	\end{gathered}
	\label{eq:moo}
\end{equation}

In MOO, no solution addresses all objective functions simultaneously. As a result, the priority should be placed on finding solutions that cannot make any goals better without making at least one of the other goals worse. Therefore, a solution $\vec{x}_1\in X$ is said to dominate another one $\vec{x}_2\in X$ if the conditions expressed in Equation \ref{eq:cons} are met.

\begin{equation}
	\begin{gathered}
	s_i(\vec{x}_1) \leq s_i(\vec{x}_2) \ \forall i \in \left\{ {1,2,3,\dots,n } \right\} \\
	s_j(\vec{x}_1) < s_j(\vec{x}_2) \ \exists j \in \left\{ {1,2,3,\dots,n } \right\}
	\end{gathered}
	\label{eq:cons}
\end{equation}

In this way, a solution $\vec{x}\in X$ is optimal if no solution might dominate it. An element $x$ is said to dominate another element $y$ if $x$ is not worse than $y$ concerning all the goals and is strictly better than $y$ for at least one. The elements of the search space that are not dominated give rise to a Pareto front, which represents the best possible solutions to the orthogonal objectives.

\subsection{Implementation}
There are several implementations for MOO strategies \cite{key-gde3,key-moead}. It is beyond the scope of this paper to consider them all. However, we will look at one of the best ones, NSGA-II \citep{key-nsga2}. This strategy is summarized in Algorithm \ref{alg:the_alg2}. Its mode of operation is based on the concepts of fronts and crowding distance.

NSGA-II adheres to the basic structure of a genetic algorithm but employs a different approach to mating and selection for survival. In the NSGA-II, the first step is to select individuals in a front-wise fashion. In this way, a situation will arise where it will be necessary to divide a front because it will not be possible for all individuals to survive. 

The solutions are chosen according to the crowding distance for this particular splitting front. Within the parameters of the objective space, the Manhattan Distance corresponds to the crowding distance. On the other hand, it is desired that the extreme points be maintained with each new generation, and as a result, an infinite crowding distance is assigned to them. Algorithm \ref{alg:the_alg2} shows us a possible implementation of this approach.

\begin{algorithm}
\caption{MOO Technique for Optimizing Text Readability}
\label{alg:the_alg2}
\begin{algorithmic}[1]
\Procedure{ORUGA2-MOO}{}
\State \textit{population $\gets$ initializePopulation ()}
\State \textit{population $\gets$ generationRandomIndividual (population)}
\State \textit{calculateReadabilityScore (population)}
\State \textit{assignRankBasedOnPareto (population)}
\State \textit{auxiliarPopulation $\gets$ generationChildPopulation (population)}
\State \textbf{while} \textit{(stop condition not reached)} \textbf{do}
\State \ \ \ \ \textbf{for} \textit{(each individual in population \textbf{and} auxiliarPopulation)} \textbf{do}
\State \ \ \ \ \ \ \textit{solution $\gets$ calculateReadabilityScore (population)}
\State \ \ \ \ \ \ \textit{solution $\gets$ assignRankBasedOnPareto (population)}
\State \ \ \ \ \ \ \textit{solution $\gets$ generateNonDominateSolutions (population)}
\State \ \ \ \ \ \ \textit{solution $\gets$ determiningCrowdingDistance (population)}
\State \ \ \ \ \ \ \textbf{for} \textit{(each solution)} \textbf{do}
\State \ \ \ \ \ \ \ \ \ \textit{population $\gets$ addingSolutionsNextGeneration (population)}
\State \ \ \ \ \ \ \textbf{end for}
\State \ \ \ \ \textbf{end for}
\State \ \ \ \ \textit{population $\gets$ selectPointsLowFrontHighCrowdingDistance (population)}
\State \ \ \ \ \textit{population $\gets$ generationNextPopulation (population)}
\State \textbf{end while}
\State optimizedText $\gets$ \textit{optimizedIndividual (solution)}
\State optimizedText $\gets$ \textit{correctErrorsIfNecessary (optimizedText)}
\State \textbf{return} \textit{optimizedText}
\EndProcedure
\end{algorithmic}
\end{algorithm}

NSGA-II is an example of a genetic algorithm, and it possesses the three characteristics listed below: It operates based on an elitist principle, which states that only the most privileged members of a population are permitted to be passed down to subsequent generations. In addition, it employs a mechanism specifically designed to preserve diversity (crowding distance). As a direct consequence of this, it can identify non-dominated solutions.

\subsection{Illustrative examples}
Since we are trying to optimize two orthogonal objectives simultaneously, it is impossible to offer a single solution. Nevertheless, we can put in the hands of the human operator a front of solutions ranging from a total optimization by modifying the most significant number of words to a minor optimization by touching a minimum number of words. It is up to the human operator to decide which solution to choose. 

Example \ref{ex:D} shows a real trace that controls the number of words to be replaced using the MOO technique known as NSGA-II. The text on which it operates is about geography and has been extracted from Wikipedia. Once again, we must insist that although, in theory, it would be possible to edit the words manually to satisfy the criteria of a given metric, this approach is transparent and works automatically for any desired metric. Please note that the words in blue are candidates to be replaced by a synonym.

\begin{table}[h]
\begin{example}[label=ex:D,colback=black!5!white,colframe=black!40!black,title=Example 4. Geography \textbf{Source:} Wikipedia]
\textbf{Goal:} \textit{Minimize the FKGL score by using Wordnet synonyms and minimize the words to be replaced using NSGA-II}. \\ \\
``Niagara Falls is a \textcolor{blue}{group} of three \textcolor{blue}{waterfalls} at the southern end of Niagara Gorge, spanning the \textcolor{blue}{border} between the \textcolor{blue}{province} of Ontario in Canada and the \textcolor{blue}{state} of New York in the United States. The largest of the three is Horseshoe Falls, which straddles the international \textcolor{blue}{border} of the two \textcolor{blue}{countries}. It is \textcolor{blue}{also} known as the Canadian Falls. The smaller American Falls and Bridal Veil Falls lie within the United States. Bridal Veil Falls is \textcolor{blue}{separated} from Horseshoe Falls by Goat Island and from American Falls by Luna Island, with both islands \textcolor{blue}{situated} in New York. Formed by the Niagara River, which drains Lake Erie into Lake Ontario, the \textcolor{blue}{combined} falls have the highest flow rate of any waterfall in North America that has a vertical drop of more than 50 m (160 ft). During peak \textcolor{blue}{daytime} tourist \textcolor{blue}{hours}, more than 168,000 m3 (5.9 million cu ft) of water goes over the crest of the falls every minute.'' \textcolor{blue}{\textbf{FKGL score: 10.72}}. 

\tcblower  

\centering
\begin{tabular}{|c|c|}
\hline
\textbf{Words to be replaced} & \textbf{FKGL expected}\\
\hline
5                    & 10.43  ($\triangledown$ 2.71\%)       \\
6                    & 10.28  ($\triangledown$ 4.10\%)       \\
7                    & 10.13  ($\triangledown$ 5.50\%)       \\
8                    & 9.98   ($\triangledown$ 6.90\%)       \\
9                    & 9.91   ($\triangledown$ 7.56\%)       \\
10                   & 9.84   ($\triangledown$ 8.21\%)       \\
11                   & 9.76   ($\triangledown$ 8.96\%)       \\
12                   & 9.68   ($\triangledown$ 9.70\%)       \\
13                   & 9.61   ($\triangledown$ 10.35\%)       \\
\hline
\end{tabular}
\end{example}
\end{table}

As it is possible to observe, minor changes can be made to the original text and still optimize readability. It is still an open question whether minor changes in form are not so minor in meaning. But that open question will be addressed later in this paper.

\subsection{Experimental study}
This section will explain the specifics of an empirical study we conducted to determine whether this novel approach is feasible. To accomplish this, we initially prepared an experimental setup so that we could determine the parameters of the experiments. Second, we completed the test and collected the unprocessed data. We have moved forward with the analysis of the data that we have gathered, which brings us to our third and last point.

\subsubsection{Experimental setup}
As was the case in the preceding part, the meticulous fine-tuning of the MOO strategy's parameters will not be the primary focus of this work. As a result, following a brief preliminary study based on a scheme of traditional parameter settings, one configuration works quite well. As a direct consequence of this, the following is a list of the parameters that we used for our experiments:

\begin{itemize}
	\item Population size \{10, 15, 20\}: \textbf{20}
	\item Number of parents mating \{10, 15, 20\}: \textbf{20}
	\item Number of genes: one per candidate word to be substituted by a synonym
	\item Fitness function: the user can choose among Equations \ref{eq:dcrf}, \ref{eq:smog}, \ref{eq:ari}, \ref{eq:fkgl}, words to be replaced
	\item Stop condition: \{300, 600, 900\}: \textbf{900} generations
\end{itemize}

Furthermore, every experiment was run on a standard computer with 32 GB of primary memory and an Intel Core i7-8700 processor running at 3.20 GHz on Microsoft Windows 10 64-bit. Most of the functionality has been implemented using the library jMetalPy\footnote{https://jmetal.github.io/jMetalPy/}, an open-source Python library for designing and implementing MOO strategies \citep{key-benitez}.

\subsubsection{Experiments}
Now we will proceed with the experiments concerning minimizing the impact on the original message's form. While we focused previously on pure optimization, we focus here on minimizing the number of replaced words to affect how the message looks as little as possible.

\begin{figure}[H]
\centering
    \begin{tikzpicture}[scale=0.5]
        \begin{axis}[title=Use Case \#1 - science, xlabel=Words to be replaced, ylabel=Readability Score, legend pos=north east]
            \addplot[magenta, line width = 1.2, mark=square] table[x=w, y=r] {nsga20.dat};
						\addlegendentry{FKGL}
		\end{axis}
    \end{tikzpicture}
		\begin{tikzpicture}[scale=0.5]
        \begin{axis}[title=Use Case \#2 - history, xlabel=Words to be replaced, ylabel=Readability Score, legend pos=north east]
						\addplot[magenta, line width =  1.2, mark=diamond] table[x=w, y=r] {nsga21.dat};
						\addlegendentry{FKGL}
		\end{axis}
    \end{tikzpicture}
		\begin{tikzpicture}[scale=0.5]
        \begin{axis}[title=Use Case \#3 - geography, xlabel=Words to be replaced, ylabel=Readability Score, legend pos=north east]
						\addplot[red, line width = 1.2, mark=square] table[x=w, y=r] {nsga22.dat};
						\addlegendentry{FKGL}
		\end{axis}
    \end{tikzpicture}
		\begin{tikzpicture}[scale=0.5]
        \begin{axis}[title=Use Case \#4 - sports, xlabel=Words to be replaced, ylabel=Readability Score, legend pos=north east]
						\addplot[black, line width =  1.2, mark=diamond] table[x=w, y=r] {nsga23.dat};
						\addlegendentry{FKGL}
		\end{axis}
    \end{tikzpicture}
    \begin{tikzpicture}[scale=0.5]
        \begin{axis}[title=Use Case \#5 - engineering, xlabel=Words to be replaced, ylabel=Readability Score, legend pos=north east]
            \addplot[black, line width = 1.2, mark=square] table[x=w, y=r] {nsga24.dat};
            \addlegendentry{FKGL}
		\end{axis}
    \end{tikzpicture}
		\begin{tikzpicture}[scale=0.5]
        \begin{axis}[title=Use Case \#6 - geography, xlabel=Words to be replaced, ylabel=Readability Score, legend pos=north east]
						\addplot[black, line width =  1.2, mark=diamond] table[x=w, y=r] {nsga25.dat};
						\addlegendentry{FKGL}
		\end{axis}
    \end{tikzpicture}
		\begin{tikzpicture}[scale=0.5]
        \begin{axis}[title=Use Case \#7 - engineering, xlabel=Words to be replaced, ylabel=Readability Score, legend pos=north east]
						\addplot[magenta, line width = 1.2, mark=square] table[x=w, y=r] {nsga26.dat};
            \addlegendentry{FKGL}
		\end{axis}
    \end{tikzpicture}
		\begin{tikzpicture}[scale=0.5]
        \begin{axis}[title=Use Case \#8 - history, xlabel=Words to be replaced, ylabel=Readability Score, legend pos=north east]
						\addplot[magenta, line width =  1.2, mark=diamond] table[x=w, y=r] {nsga27.dat};
						\addlegendentry{FKGL}
		\end{axis}
    \end{tikzpicture}
    \begin{tikzpicture}[scale=0.5]
        \begin{axis}[title=Use Case \#9 - sports, xlabel=Words to be replaced, ylabel=Readability Score, legend pos=north east]
            \addplot[red, line width = 1.2, mark=square] table[x=w, y=r] {nsga28.dat};
            \addlegendentry{FKGL}
		\end{axis}
    \end{tikzpicture}
		\begin{tikzpicture}[scale=0.5]
        \begin{axis}[title=Use Case \#10 - history, xlabel=Words to be replaced, ylabel=Readability Score, legend pos=north east]
						\addplot[blue, line width =  1.2, mark=diamond] table[x=w, y=r] {nsga29.dat};
						\addlegendentry{FKGL}
		\end{axis}
    \end{tikzpicture}
	\caption{Non-dominated solutions for ten use cases obtained using NSGA-II}
	\label{fig:p2}
\end{figure}
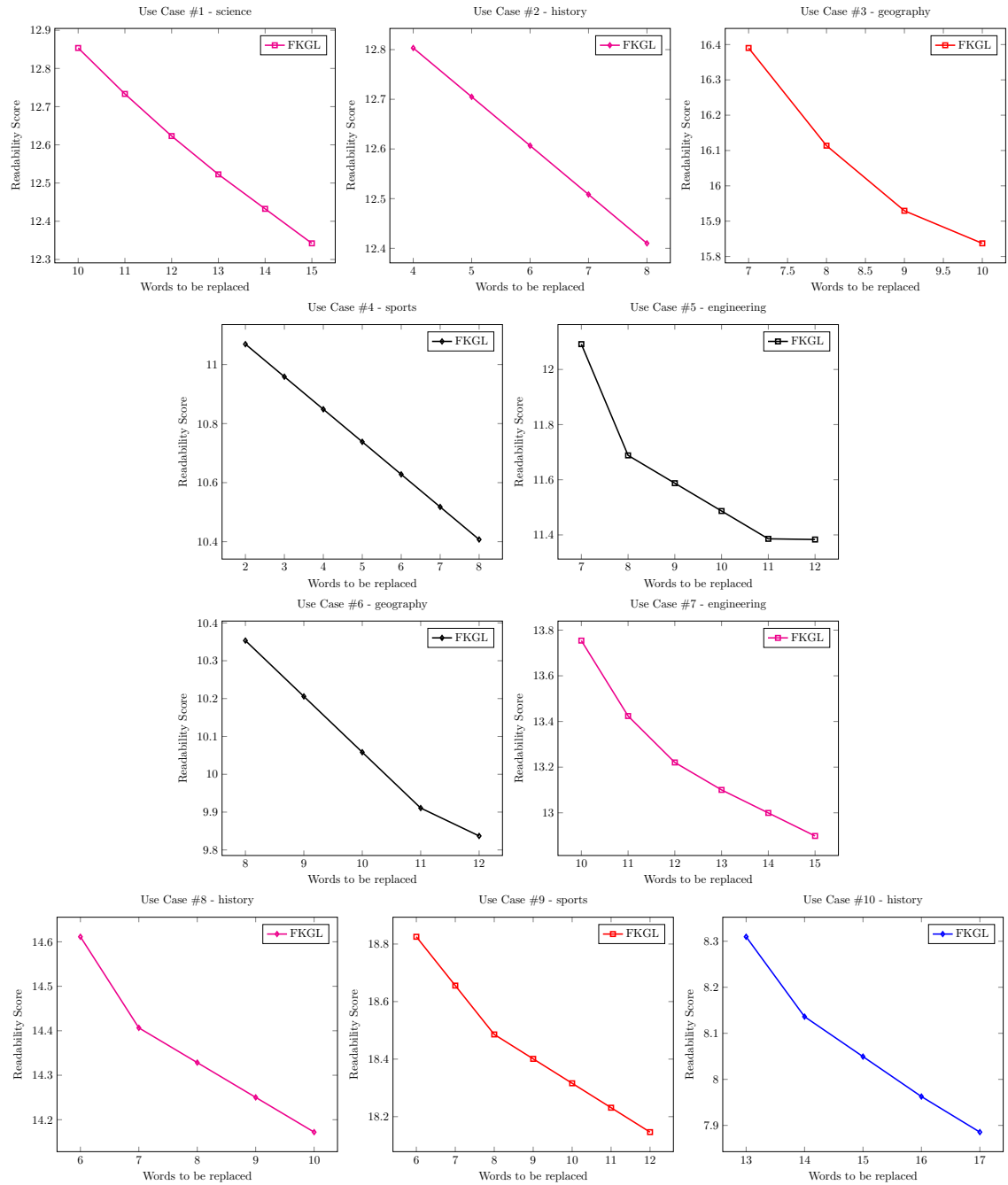

In Figure \ref{fig:p2}, we can see a summary of the results obtained after our experiments. What we have done is try to minimize the readability score (FKGL) at the same time as the number of words to be replaced in the ten use cases we are using throughout this work. As can be seen, we always obtain a Pareto front of solutions which indicates that the fewer words replaced, the less optimization is carried out. However, it should also be noted that the fewer words replaced also means a more negligible impact on the form of the initial message. It should be noted that different colors have been used for the Pareto fronts as in the previous experiments. Each color represents the degree of difficulty of each case study.

\newpage

\subsubsection{Discussion}
We have shown how it is possible to design a solution that enhances the readability of texts, and we implemented that solution through MOO techniques. In addition, this sort of optimization is carried out while paying attention to the form of the texts in question to minimize the impact of replacing words with synonyms that better fit the readability criteria of the various formulas we have researched.

As we explained earlier the paper, this kind of optimization can also occur in two distinct ways: it can be carried out to lower the readability score, making it possible for more people to comprehend the text altogether. Also, as a second possibility, it is possible to do so in order to improve the readability score, which will enable a lesser number of individuals to comprehend the processed text.

Although the solutions will always be presented as a Pareto front for the human operator to choose his solution, it is also possible to define profiles: conservative, medium, and aggressive, which will opt for the conservative, medium, or more daring versions respectively without any question, without prejudice to the fact that it should be possible to manually edit a word that does not fit the context.

\section{Part III: Preserving the essence of the original text}
\label{sec:partIII}
In Section \ref{sec:partI}, we have seen how it is possible to design and implement a functional solution to optimize the readability of any text. Although such a solution seems to work quite well, it cannot be considered fully automatic since a human user still has to evaluate whether any word used to replace original terms is not out of place in the context of the original message. In Section \ref{sec:partII}, we have used a MOO strategy to ensure that we only replace a minimum number of words in the original text in order to preserve its form, and we even go a step further and allow a human operator to decide the degree of impact of ORUGA on the form of the original text. In Section \ref{sec:partIII}, we will address a remaining problem, which is to keep under control the semantic distance between the original text and the generated text so that a short distance guarantees that the original text has not been distorted, while a long-distance means that we have been able to optimize readability by a large amount, but at a high cost in terms of altering the meaning of the original text. The importance of this third and final part is that success in our strategy is what can guarantee that the solution is entirely unsupervised.

\subsection{Technical Preliminaries}
The additional element we will add in this part is to measure the semantic distance between the original text and the text to be delivered. In this way, we can control any significant change in the meaning of the message. This semantic distance will be the third objective in our MOO strategy.

In the literature, there are methods that allow us to determine the semantic distance between two pieces of text in a meaningful way, even when those pieces do not share any words. To do that, words are embedded as vectors using this method. It has been demonstrated to perform better than many of the methods considered to be state-of-the-art in the k-nearest neighbor's classification.

With the help of Word Mover's Distance (WMD) \citep{key-kusner}, and given pre-trained word embeddings, it is possible to automatically assess the semantic distance between two texts by computing the minimum distance that the embedded words of one text need to travel to reach the ones of another text. So, for example, we can use the WordNet library to calculate synonyms that allow us to optimize text readability. At the same time, we can use word2vec to supervise that the semantic between the generated text and the original text is minimized. We could use as a metaphor that it is an adversarial process.

The beauty of this approach is that the different synonym libraries now do not compete with each other but collaborate to try to measure (and therefore facilitate control) the semantic distance between the original text and the final text. Values close to zero will indicate that the meaning of the texts under consideration is practically equivalent, while distances approaching infinity indicate that the texts are incredibly different. 

Moreover, we do not have to concern about whether the genetic algorithm replaces a word with one or more words (e.g., 'considering' by 'taking into account') since the WMD is prepared for this contiguity, as it assumes by design that the texts will not have the same length. Since when using WMD, each word is matched against all other words, but weighted by a flow matrix $T$ that ensures the semantic distance will be symmetric, even when an unequal number of words must be matched.

\subsection{Implementation}
In this work, we have decided to use WMD to facilitate the measurement of the semantic distance between the initial text and the text that will be delivered at the end of the process. This ensures that the synonyms used to replace candidate words are coherent. This choice is because WMD can measure the amount of semantic distance that separates two pieces of text by comparing the words that are important to each other. This is true even if the two pieces of text do not share any words.

In addition to that, the method makes use of a representation known as the bag-of-words representation. The idea behind this method is that it should be possible to figure out how far apart two different texts are by figuring out the best way to move the distribution of the source text and the text being targeted. 

We can formally define our strategy, so that let $d$ and $d'$ be the embedding representation of two texts, and $T \in \mathcal{R}^{n \times n}$ where $T_{ij} \geq 0$ means how much of word $i$ in $d$ travels to word $j$ in $d'$. Furthermore, the distance between $i$ and $j$ might be $c(i, j) = \left\|x_{i} - x_{j}\right\|$. By $c(i, j)$, we denote the cost of moving from one word to another. In order to transform $d$ into $d'$, it is necessary to be sure that the flow from $i$ is equivalent to $d_i$ so that $\sum_{j}T_{ij} = d_i$. In this way, the minimum cumulative cost of moving $d$ to $d'$, given all these constraints, is provided by the solution shown in Equation \ref{eq:wmd}.

\begin{equation}
	  \begin{gathered}
	arg \ min \			\sum_{i,j=1}^{n}T_{ij}c(i, j) \\
	subject \ to \ 	\sum_{j=1}^{n}T_{ij} = d_i \ \forall i \in \{1, 2, 3 \cdots n\} \ \wedge \\
			\ \ \ \			\sum_{i=1}^{n}T_{ij} = d'_j \ \forall j \in \{1, 2, 3 \cdots n\}
  \end{gathered}
	\label{eq:wmd}
\end{equation}

Therefore, we use a function that determines the distance between two texts as the cumulative sum of the minimum distance each word in one text must move in vector space to the closest word in the other text. Word embeddings derived from word2vec will be utilized for this work because of their capability to maintain critical aspects of the context in which a word is used. 

WMD is used quite frequently these days to calculate semantic distances, and this is one of the reasons why. The only problem is that the complexity of computing the constrained minimum cumulative cost in the worst case is $O(p^3 \log u)$, where $u$ is the number of unique words in the text \citep{skianis2020boosting}. Therefore, when working with texts that contain a large number of unique words, WMD may perform poorly. However, there are some techniques that improve performance.

In the context of this work, we have used the library \textit{gensim}\footnote{https://pypi.org/project/gensim/} implementation of the WMD fed by the word embeddings from word2vec \citep{key-Mikolov}. In this way, what was previously a rival synonym library now becomes an adversarial library that helps keep semantic distance under control.

\subsection{Illustrative examples}
Example \ref{ex:E} provides us with information about engineering that was taken from Wikipedia. We are interested in observing how ORUGA operates while attempting to minimize the FKGL readability score, the number of words that need to be replaced, and the semantic distance between the original text and the text that has been processed. We are replacing the synonyms with the help of WordNet, and we are moving forward with the MOO with the help of NSGA-II. The ultimate goal is to ensure a very low risk of distortion of the original message that wanted to be communicated.

\begin{table}[H]
\begin{example}[label=ex:E,colback=black!5!white,colframe=black!40!black,title=Example 5. Engineering]
\begin{tcolorbox}[colback=black!10!white,colframe=black!70!black,title=Original text \textbf{Source:} Wikipedia,colbacktitle=black!50!white]
``Big data refers to data sets that are too large or complex to be dealt with by traditional data-processing application software. Data with many fields (rows) offer greater statistical power, while data with higher complexity (more attributes or columns) may lead to a higher false discovery rate. Big data analysis challenges include capturing data, data storage, data analysis, search, sharing, transfer, visualization, querying, updating, information privacy, and data source. Big data was originally associated with three key concepts volume, variety, and velocity. The analysis of big data presents challenges in sampling, and thus previously allowing for only observations and sampling. Thus a fourth concept, veracity, refers to the quality or insightfulness of the data.'' \textcolor{blue}{\textbf{ARI score: 14.43}}.
\end{tcolorbox}

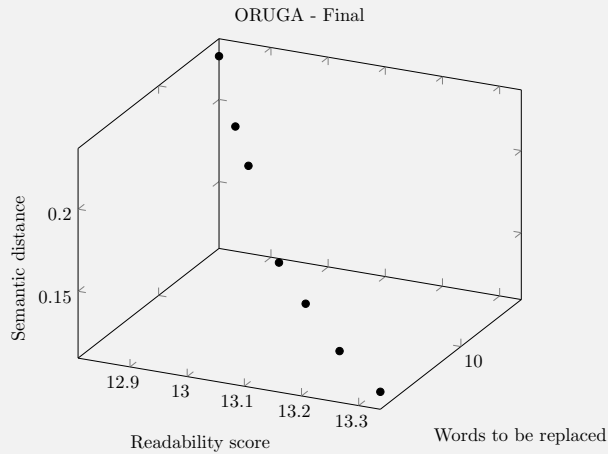
\begin{tcolorbox}
\begin{center}
	\begin{tikzpicture}[scale=0.7]
	\begin{axis}[title=ORUGA - Final, xlabel=Readability score, ylabel=Words to be replaced, zlabel=Semantic distance, legend pos=outer north east]
 
\addplot3[black, only marks] coordinates {
(13.337894736842104, 6, 0.11993842287937737)
(13.231315789473683, 7, 0.12891337784993515)
(13.136811594202896, 8, 0.14263657417805922)
(13.054898550724637, 9, 0.15330857281710278)
(12.931072463768118, 11, 0.18584001591682167) 
(12.872985507246378, 12, 0.19680782596398577) 
(12.809159420289859, 13, 0.22631948591388285)
};
 
\end{axis}
\end{tikzpicture}
\end{center}
\end{tcolorbox}

\begin{tcolorbox}[colback=black!10!white,colframe=black!70!black,title=Minimize ARI score with the least risk of distorting the original message,colbacktitle=black!50!white]
Big data \textcolor{blue}{bring up} to data sets that are too \textcolor{blue}{big} or complex to be dealt with by traditional data-processing application software. Data with many fields (rows) offer greater statistical power, while data with higher complexity (more attributes or columns) may lead to a higher false \textcolor{blue}{finding} rate. Big data analysis challenges include capturing data, data storage, data analysis, search, sharing, transfer, visualization, querying, updating, information privacy, and data source. Big data was originally \textcolor{blue}{tied in} with three key concepts volume, variety, and velocity. The analysis of big data presents challenges in sampling, and thus previously \textcolor{blue}{let} for only observations and sampling. Thus a fourth concept, veracity, refers to the quality or \textcolor{blue}{acumen} of the data. \textcolor{blue}{\textbf{ARI score: 13.33 - $\triangledown$ 7.62\%}}
\end{tcolorbox}

\end{example}
\end{table}

The example shows that we no longer operate as blindly as before. Now, we also minimize the words to be replaced and the semantic distance between the initial and the generated text. Therefore, the results are much more reasonable and can be relied upon to work in exploitation environments. The readability score optimization is less spectacular than before, but the risk of distorting the original message, both in form and content, is much more substantially reduced.

\subsection{Experimental study}
In this section, we have performed the empirical study to test this version of ORUGA. To do so, we designed the experiments through an experimental setup. We performed the experiments and collected the raw data. And finally, we proceeded with the analysis of the collected data.

\subsubsection{Experimental setup}
The accurate adjustment of the MOO strategy's parameters is not the primary focus here, as was the case with the previous parts of this research. One configuration works quite well after a brief preliminary study based on a conventional parameter-setting strategy. This came about as a result of what was mentioned above. The following is a list of the parameters that we have used:

\begin{itemize}
	\item Population size \{10, 15, 20\}: \textbf{20}
	\item Number of parents mating \{10, 15, 20\}: \textbf{20}
	\item Number of genes: one per candidate word to be substituted by a synonym
	\item Fitness function: Equations \ref{eq:dcrf}, \ref{eq:smog}, \ref{eq:ari}, \ref{eq:fkgl}, words to be replaced, and WMD \citep{key-kusner}.
	\item Stop condition: \{300, 600, 900\}: \textbf{900} generations
\end{itemize}

\subsubsection{Experiments}
We are going to proceed with the experiments concerning minimizing the impact on the meaning of the original message. We focus now on being able to produce a result that can be put (or be close to being put into exploitation). To do this, we will conduct experiments to see if we can control the difference in meaning between the original message's content and the generated text.

\begin{figure}[H]
	\centering
	\begin{tikzpicture}[scale=0.45]
	\begin{axis}[title=Use Case \#1 - science, xlabel=Readabity score, ylabel=Words to be replaced, zlabel=Semantic distance, legend pos=outer north east]
 \addplot3[magenta, only marks] coordinates {
(13.50562264,	7,	0.070884348)
(13.27471028,	8,	0.086390007)
(13.39430189,	8,	0.080881085)
(13.28298113,	8,	0.085192427)
(13.05414953,	9,	0.107522428)
(13.16442991,	9,	0.096405124)
(13.17166038,	9,	0.094756856)
(13.06033962,	10,	0.106422459)
(12.94386916,	10,	0.11298982)
(12.83358879,	11,	0.122701342)
(12.83096296,	11,	0.129723602)
(12.72330841,	12,	0.138093167)
(12.61302804,	13,	0.155301309)
(12.61244444,	13,	0.157011267)
(12.50318519,	14,	0.174263462)
};
\end{axis}
\end{tikzpicture}
\begin{tikzpicture}[scale=0.45]
	\begin{axis}[title=Use Case \#2 - history, xlabel=Readabity score, ylabel=Words to be replaced, zlabel=Semantic distance, legend pos=outer north east]
 \addplot3[magenta, only marks] coordinates {
(13.09833333,	1,	0.008054845)
(12.90166667,	1,	0.016714507)
(12.80333333,	2,	0.022803502)
(13,	2,	0.016628288)
(12.705,	2,	0.031247512)
(12.60666667,	3,	0.036842483)
(12.705,	3,	0.029334291)
(12.50833333,	4,	0.042999127)
(12.41,	5,	0.049979719)
(12.31166667,	6,	0.058859616)

};
\end{axis}
\end{tikzpicture}
	\begin{tikzpicture}[scale=0.45]
	\begin{axis}[title=Use Case \#3 - geography, xlabel=Readabity score, ylabel=Words to be replaced, zlabel=Semantic distance, legend pos=outer north east]
 \addplot3[red, only marks] coordinates {
(16.7590625,	5,	0.054344518)
(16.5746875,	6,	0.064441682)
(16.3903125,	7,	0.07644426)
(16.298125,	8,	0.089846122)
(16.2059375,	8,	0.090906398)
(16.11375,	9,	0.104264999)
(16.0215625,	9,	0.106169995)
(15.929375,	10,	0.119371424)
(15.8371875,	11,	0.133461316)
(15.745,	12,	0.151573135)
};
\end{axis}
\end{tikzpicture}
\begin{tikzpicture}[scale=0.45]
	\begin{axis}[title=Use Case \#4 - sports, xlabel=Readabity score, ylabel=Words to be replaced, zlabel=Semantic distance, legend pos=outer north east]
 \addplot3[black, only marks] coordinates {
(10.95882243,	4,	0.046359696)
(10.84854206,	5,	0.053626347)
(10.73826168,	6,	0.062919182)
(10.62798131,	7,	0.072910759)
(10.51770093,	8,	0.088340143)
(10.40742056,	9,	0.105346016)

};
\end{axis}
\end{tikzpicture}
	\begin{tikzpicture}[scale=0.45]
	\begin{axis}[title=Use Case \#5 - engineering, xlabel=Readabity score, ylabel=Words to be replaced, zlabel=Semantic distance, legend pos=outer north east]
 \addplot3[black, only marks] coordinates {
(11.99977011,	6,	0.090446892)
(11.89804598,	7,	0.100509354)
(11.6945977,	8,	0.105400116)
(11.48666667,	8,	0.124165771)
(11.59287356,	9,	0.115220844)
(11.38581197,	9,	0.133022719)
(11.28495726,	10,	0.142492211)
(11.18410256,	11,	0.155362228)
(11.08324786,	12,	0.165661931)
(10.98239316,	13,	0.175169433)
(10.88153846,	14,	0.189505034)
(10.78068376,	15,	0.202939462)
};
\end{axis}
\end{tikzpicture}
\begin{tikzpicture}[scale=0.45]
	\begin{axis}[title=Use Case \#6 - geography, xlabel=Readabity score, ylabel=Words to be replaced, zlabel=Semantic distance, legend pos=outer north east]
 \addplot3[black, only marks] coordinates {
(10.13196429,	5,	0.050543261)
(10.05821429,	6,	0.058381343)
(9.984464286,	6,	0.066105817)
(9.910714286,	7,	0.073724405)
(9.836964286,	8,	0.082882668)
(9.763214286,	9,	0.092978913)
(9.689464286,	10,	0.101350347)
(9.615714286,	11,	0.111599056)
(9.541964286,	12,	0.125440735)
};
\end{axis}
\end{tikzpicture}
	\begin{tikzpicture}[scale=0.45]
	\begin{axis}[title=Use Case \#7 - engineering, xlabel=Readabity score, ylabel=Words to be replaced, zlabel=Semantic distance, legend pos=outer north east]
 \addplot3[magenta, only marks] coordinates {
(13.47035088,	15,	0.16383418)
(13.34376812,	16,	0.177048266)
(13.22045977,	17,	0.191404398)
(13.11873563,	18,	0.209625159)
(13.01701149,	19,	0.232023701)
};
\end{axis}
\end{tikzpicture}
\begin{tikzpicture}[scale=0.45]
	\begin{axis}[title=Use Case \#8 - history, xlabel=Readabity score, ylabel=Words to be replaced, zlabel=Semantic distance, legend pos=outer north east]
 \addplot3[magenta, only marks] coordinates {
(14.32838411,	8,	0.080626494)
(14.25023841,	9,	0.092332707)
(14.17209272,	10,	0.106876612)
(14.0744,	11,	0.117664)
};
\end{axis}
\end{tikzpicture}
	\begin{tikzpicture}[scale=0.45]
	\begin{axis}[title=Use Case \#9 - sports, xlabel=Readabity score, ylabel=Words to be replaced, zlabel=Semantic distance, legend pos=outer north east]
 \addplot3[red, only marks] coordinates {
(19.07997602,	5,	0.065100875)
(18.82529976,	6,	0.077060269)
(18.99508393,	6,	0.072357953)
(18.74040767,	7,	0.083811789)
(18.65551559,	7,	0.091635579)
(18.65551559,	8,	0.091516945)
(18.5706235,	8,	0.100576286)
(18.5706235,	9,	0.100432154)
(18.48573141,	9,	0.108149947)
(18.40083933,	10,	0.117366389)
(18.31594724,	11,	0.131088184)
(18.23105516,	12,	0.144984542)
};
\end{axis}
\end{tikzpicture}
\begin{tikzpicture}[scale=0.45]
	\begin{axis}[title=Use Case \#10 - history, xlabel=Readabity score, ylabel=Words to be replaced, zlabel=Semantic distance, legend pos=outer north east]
 \addplot3[blue, only marks] coordinates {
(8.316040146,	11,	0.10336417)
(8.143777372,	12,	0.105374419)
(8.057645985,	13,	0.117085325)
(7.971514599,	15,	0.139401641)
};
\end{axis}
\end{tikzpicture}
\caption {Summary of the results obtained for the third (and final) version of ORUGA}
\label{fig:final}
\end{figure}
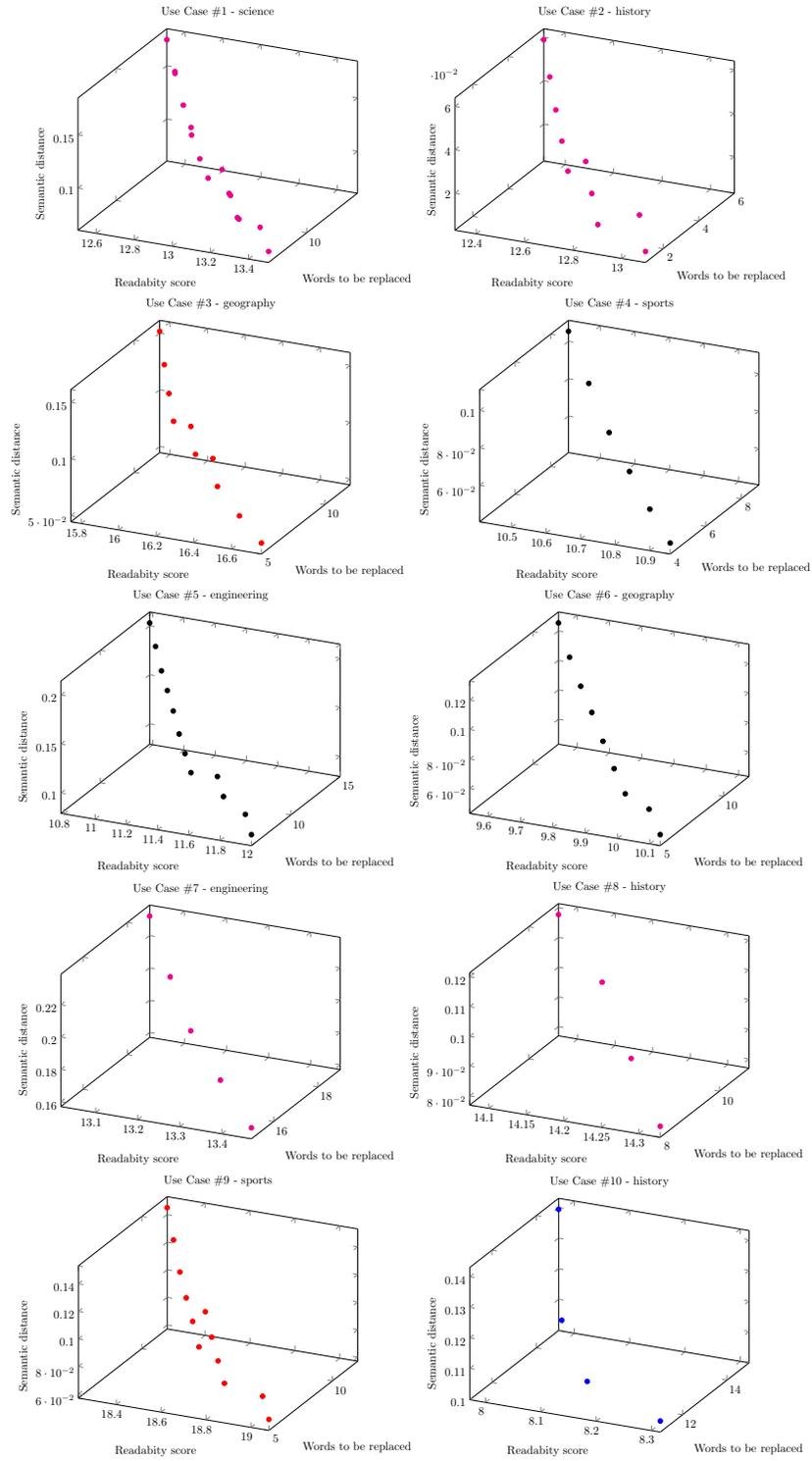

Figure \ref{fig:final} shows us the summary of all the results obtained for the ten use cases that we have been studying throughout this research work. As can be seen, each use case, no matter how topical or challenging, corresponds to a good number of solutions ranging from the most conservative (the one that has the least risk of distorting the original message) to the most aggressive (the one that reduces the readability score more conclusively at the risk of distorting the original message). According to previous experiments, each color represents a degree of difficulty. 

\pagebreak

\subsubsection{Discussion}
We have seen how it is possible to build a solution that enhances the readability of texts of very different natures and readability levels, and we implemented that solution. In addition, this sort of optimization is conducted while paying attention to both the content and the structure of the texts in question to minimize the impact of replacing words with synonyms that are a better fit for the readability criteria of the various formulas we have researched.

Throughout this research, we have seen how it is possible to use genetic algorithms to improve the readability of any text formulated in English. As we have explained earlier, optimization is bi-directional. That is, it can automatically increase or decrease the readability of the text. Once we established the basis for the optimization, we could control the number of words that could be replaced; we could also control that the semantic distance between the original text and the final text does not skyrocket. Therefore, our initial goals of building a solution that improves the readability of any text without significantly altering its form or content have been achieved.

\section{Conclusion}
In this research work, we have seen how ORUGA can automatically optimize the readability of a text by using genetic algorithms. We have shown that by automatically replacing some words of the text to be optimized by their synonyms, we can optimize the readability levels in the direction (minimize or maximize) we wish. Neither the content nor the form of the text is altered because a minimal impact on the transformation of the original text is sought through various additional MOO techniques. Although, in theory, analogous solutions could be built using neural language models, this approach has a significant advantage: it is unsupervised and requires no training. 

An exhaustive empirical study has shown that we have successfully performed all experiments. In the first instance, these experiments consisted of processing texts of different natures (history, geography, sports, nature, and science) using three synonym libraries and using different readability metrics. In the second instance: designing a MOO solution to control the number of words to be replaced. We have tested different texts to assess and compare the feasibility of this approach. Furthermore, in the third instance: measuring and controlling the semantic distance between the original text and the one that will finally be outputted. For this, we have used a novel technique that uses a library of synonyms to control the results obtained with another library of different synonyms in an adversarial process. 

The results demonstrate that our hypothesis about text readability optimization at the beginning of this paper is valid. Despite the success of this research, it is necessary to bear in mind that simple formulas are typically simpler to put into practice, which is a limitation of this body of work. These formulas have a fundamental inability to model the semantics of word usage in context, which is needed to capture richer ideas of text difficulty. 

As future work, a possible approach (yet computationally expensive) would be using a model of contextual embeddings such as BERT on a large dataset. Then, it should be necessary to store pairs of words and corresponding contextual representations and then use the nearest neighbors approach to identify synonyms that include the context. In this way, the impact of text processing will be even more limited.

\section*{Source code}
The source code of this approach is published under MIT license in the following Github repository: \url{https://github.com/jorge-martinez-gil/oruga}. 

\section*{Acknowledgments}
The research reported in this paper has been funded by the Federal Ministry for Climate Action, Environment, Energy, Mobility, Innovation, and Technology (BMK), the Federal Ministry for Digital and Economic Affairs (BMDW), and the State of Upper Austria in the frame of SCCH, a center in the COMET - Competence Centers for Excellent Technologies Programme managed by Austrian Research Promotion Agency FFG.

%

\bibliography{mybib}

\begin{thebibliography}{36}
\expandafter\ifx\csname natexlab\endcsname\relax\def\natexlab#1{#1}\fi
\providecommand{\url}[1]{\texttt{#1}}
\providecommand{\href}[2]{#2}
\providecommand{\path}[1]{#1}
\providecommand{\DOIprefix}{doi:}
\providecommand{\ArXivprefix}{arXiv:}
\providecommand{\URLprefix}{URL: }
\providecommand{\Pubmedprefix}{pmid:}
\providecommand{\doi}[1]{\href{http://dx.doi.org/#1}{\path{#1}}}
\providecommand{\Pubmed}[1]{\href{pmid:#1}{\path{#1}}}
\providecommand{\bibinfo}[2]{#2}
\ifx\xfnm\relax \def\xfnm[#1]{\unskip,\space#1}\fi
\bibitem[{Ante(2022)}]{key-Ante}
\bibinfo{author}{Ante, L.} (\bibinfo{year}{2022}).
\newblock \bibinfo{title}{The relationship between readability and scientific
  impact: Evidence from emerging technology discourses}.
\newblock {\it \bibinfo{journal}{J. Informetrics}\/},  {\it
  \bibinfo{volume}{16}\/}, \bibinfo{pages}{101252}. \URLprefix
  \url{https://doi.org/10.1016/j.joi.2022.101252}.
  \DOIprefix\doi{10.1016/j.joi.2022.101252}.
\bibitem[{Benítez-Hidalgo et~al.(2019)Benítez-Hidalgo, Nebro, García-Nieto,
  Oregi \& Ser}]{key-benitez}
\bibinfo{author}{Benítez-Hidalgo, A.}, \bibinfo{author}{Nebro, A.~J.},
  \bibinfo{author}{García-Nieto, J.}, \bibinfo{author}{Oregi, I.}, \&
  \bibinfo{author}{Ser, J.~D.} (\bibinfo{year}{2019}).
\newblock \bibinfo{title}{jmetalpy: A python framework for multi-objective
  optimization with metaheuristics}.
\newblock {\it \bibinfo{journal}{Swarm and Evolutionary Computation}\/},  (p.
  \bibinfo{pages}{100598}). \URLprefix
  \url{http://www.sciencedirect.com/science/article/pii/S2210650219301397}.
  \DOIprefix\doi{https://doi.org/10.1016/j.swevo.2019.100598}.
\bibitem[{Chall \& Dale(1995)}]{chall1995readability}
\bibinfo{author}{Chall, J.~S.}, \& \bibinfo{author}{Dale, E.}
  (\bibinfo{year}{1995}).
\newblock {\it \bibinfo{title}{Readability revisited: The new Dale-Chall
  readability formula}\/}.
\newblock \bibinfo{publisher}{Brookline Books}.
\bibitem[{Chandrasekaran \& Mago(2021)}]{key-chandrasekaran}
\bibinfo{author}{Chandrasekaran, D.}, \& \bibinfo{author}{Mago, V.}
  (\bibinfo{year}{2021}).
\newblock \bibinfo{title}{Evolution of semantic similarity - {A} survey}.
\newblock {\it \bibinfo{journal}{{ACM} Comput. Surv.}\/},  {\it
  \bibinfo{volume}{54}\/}, \bibinfo{pages}{41:1--41:37}.
  \DOIprefix\doi{10.1145/3440755}.
\bibitem[{Collins-Thompson(2014)}]{collins2014computational}
\bibinfo{author}{Collins-Thompson, K.} (\bibinfo{year}{2014}).
\newblock \bibinfo{title}{Computational assessment of text readability: A
  survey of current and future research}.
\newblock {\it \bibinfo{journal}{ITL-International Journal of Applied
  Linguistics}\/},  {\it \bibinfo{volume}{165}\/}, \bibinfo{pages}{97--135}.
\bibitem[{Dale \& Chall(1948)}]{dale1948formula}
\bibinfo{author}{Dale, E.}, \& \bibinfo{author}{Chall, J.~S.}
  (\bibinfo{year}{1948}).
\newblock \bibinfo{title}{A formula for predicting readability: Instructions}.
\newblock {\it \bibinfo{journal}{Educational research bulletin}\/},  (pp.
  \bibinfo{pages}{37--54}).
\bibitem[{Deb et~al.(2002)Deb, Agrawal, Pratap \& Meyarivan}]{key-nsga2}
\bibinfo{author}{Deb, K.}, \bibinfo{author}{Agrawal, S.},
  \bibinfo{author}{Pratap, A.}, \& \bibinfo{author}{Meyarivan, T.}
  (\bibinfo{year}{2002}).
\newblock \bibinfo{title}{A fast and elitist multiobjective genetic algorithm:
  {NSGA-II}}.
\newblock {\it \bibinfo{journal}{{IEEE} Trans. Evol. Comput.}\/},  {\it
  \bibinfo{volume}{6}\/}, \bibinfo{pages}{182--197}. \URLprefix
  \url{https://doi.org/10.1109/4235.996017}.
  \DOIprefix\doi{10.1109/4235.996017}.
\bibitem[{Devlin et~al.(2019)Devlin, Chang, Lee \& Toutanova}]{key-Bert}
\bibinfo{author}{Devlin, J.}, \bibinfo{author}{Chang, M.},
  \bibinfo{author}{Lee, K.}, \& \bibinfo{author}{Toutanova, K.}
  (\bibinfo{year}{2019}).
\newblock \bibinfo{title}{{BERT:} pre-training of deep bidirectional
  transformers for language understanding}.
\newblock In \bibinfo{editor}{J.~Burstein}, \bibinfo{editor}{C.~Doran}, \&
  \bibinfo{editor}{T.~Solorio} (Eds.), {\it \bibinfo{booktitle}{Proceedings of
  the 2019 Conference of the North American Chapter of the Association for
  Computational Linguistics: Human Language Technologies, {NAACL-HLT} 2019,
  Minneapolis, MN, USA, June 2-7, 2019, Volume 1 (Long and Short Papers)}\/}
  (pp. \bibinfo{pages}{4171--4186}).
\newblock \bibinfo{publisher}{Association for Computational Linguistics}.
\newblock \DOIprefix\doi{10.18653/v1/n19-1423}.
\bibitem[{Forrest(1996)}]{key-forrest}
\bibinfo{author}{Forrest, S.} (\bibinfo{year}{1996}).
\newblock \bibinfo{title}{Genetic algorithms}.
\newblock {\it \bibinfo{journal}{ACM Computing Surveys (CSUR)}\/},  {\it
  \bibinfo{volume}{28}\/}, \bibinfo{pages}{77--80}.
\bibitem[{Kincaid et~al.(1975)Kincaid, Fishburne~Jr, Rogers \&
  Chissom}]{kincaid1975derivation}
\bibinfo{author}{Kincaid, J.~P.}, \bibinfo{author}{Fishburne~Jr, R.~P.},
  \bibinfo{author}{Rogers, R.~L.}, \& \bibinfo{author}{Chissom, B.~S.}
  (\bibinfo{year}{1975}).
\newblock {\it \bibinfo{title}{Derivation of new readability formulas
  (automated readability index, fog count and flesch reading ease formula) for
  navy enlisted personnel}\/}.
\newblock \bibinfo{type}{Technical Report} Naval Technical Training Command
  Millington TN Research Branch.
\bibitem[{Kukkonen \& Lampinen(2005)}]{key-gde3}
\bibinfo{author}{Kukkonen, S.}, \& \bibinfo{author}{Lampinen, J.}
  (\bibinfo{year}{2005}).
\newblock \bibinfo{title}{Gde3: The third evolution step of generalized
  differential evolution}.
\newblock In {\it \bibinfo{booktitle}{2005 IEEE congress on evolutionary
  computation}\/} (pp. \bibinfo{pages}{443--450}).
\newblock \bibinfo{organization}{IEEE} volume~\bibinfo{volume}{1}.
\bibitem[{Kusner et~al.(2015)Kusner, Sun, Kolkin \& Weinberger}]{key-kusner}
\bibinfo{author}{Kusner, M.}, \bibinfo{author}{Sun, Y.},
  \bibinfo{author}{Kolkin, N.}, \& \bibinfo{author}{Weinberger, K.}
  (\bibinfo{year}{2015}).
\newblock \bibinfo{title}{From word embeddings to document distances}.
\newblock In {\it \bibinfo{booktitle}{International conference on machine
  learning}\/} (pp. \bibinfo{pages}{957--966}).
\newblock \bibinfo{organization}{PMLR}.
\bibitem[{Madrazo~Azpiazu \& Pera(2020)}]{madrazo2020cross}
\bibinfo{author}{Madrazo~Azpiazu, I.}, \& \bibinfo{author}{Pera, M.~S.}
  (\bibinfo{year}{2020}).
\newblock \bibinfo{title}{Is cross-lingual readability assessment possible?}
\newblock {\it \bibinfo{journal}{Journal of the Association for Information
  Science and Technology}\/},  {\it \bibinfo{volume}{71}\/},
  \bibinfo{pages}{644--656}.
\bibitem[{Maqsood et~al.(2022)Maqsood, Shahid, Afzal, Roman, Khan, Nawaz \&
  Aziz}]{key-Maqsood}
\bibinfo{author}{Maqsood, S.}, \bibinfo{author}{Shahid, A.},
  \bibinfo{author}{Afzal, M.~T.}, \bibinfo{author}{Roman, M.},
  \bibinfo{author}{Khan, Z.}, \bibinfo{author}{Nawaz, Z.}, \&
  \bibinfo{author}{Aziz, M.~H.} (\bibinfo{year}{2022}).
\newblock \bibinfo{title}{Assessing english language sentences readability
  using machine learning models}.
\newblock {\it \bibinfo{journal}{PeerJ Comput. Sci.}\/},  {\it
  \bibinfo{volume}{8}\/}, \bibinfo{pages}{e818}. \URLprefix
  \url{https://doi.org/10.7717/peerj-cs.818}.
  \DOIprefix\doi{10.7717/peerj-cs.818}.
\bibitem[{Martinc et~al.(2021)Martinc, Pollak \&
  Robnik-{\v{S}}ikonja}]{martinc2021supervised}
\bibinfo{author}{Martinc, M.}, \bibinfo{author}{Pollak, S.}, \&
  \bibinfo{author}{Robnik-{\v{S}}ikonja, M.} (\bibinfo{year}{2021}).
\newblock \bibinfo{title}{Supervised and unsupervised neural approaches to text
  readability}.
\newblock {\it \bibinfo{journal}{Computational Linguistics}\/},  {\it
  \bibinfo{volume}{47}\/}, \bibinfo{pages}{141--179}.
\bibitem[{Martinez-Gil(2022)}]{key-martinez-mlwa}
\bibinfo{author}{Martinez-Gil, J.} (\bibinfo{year}{2022}).
\newblock \bibinfo{title}{A comprehensive review of stacking methods for
  semantic similarity measurement}.
\newblock {\it \bibinfo{journal}{Machine Learning with Applications}\/},  {\it
  \bibinfo{volume}{10}\/}, \bibinfo{pages}{100423}.
\bibitem[{Martinez-Gil \& Chaves{-}Gonzalez(2019)}]{key-martinez-eswa}
\bibinfo{author}{Martinez-Gil, J.}, \& \bibinfo{author}{Chaves{-}Gonzalez,
  J.~M.} (\bibinfo{year}{2019}).
\newblock \bibinfo{title}{Automatic design of semantic similarity controllers
  based on fuzzy logics}.
\newblock {\it \bibinfo{journal}{Expert Syst. Appl.}\/},  {\it
  \bibinfo{volume}{131}\/}, \bibinfo{pages}{45--59}.
  \DOIprefix\doi{10.1016/j.eswa.2019.04.046}.
\bibitem[{Martinez-Gil \& Chaves-Gonzalez(2020)}]{key-martinez-eswa2}
\bibinfo{author}{Martinez-Gil, J.}, \& \bibinfo{author}{Chaves-Gonzalez, J.~M.}
  (\bibinfo{year}{2020}).
\newblock \bibinfo{title}{A novel method based on symbolic regression for
  interpretable semantic similarity measurement}.
\newblock {\it \bibinfo{journal}{Expert Syst. Appl.}\/},  {\it
  \bibinfo{volume}{160}\/}, \bibinfo{pages}{113663}.
  \DOIprefix\doi{10.1016/j.eswa.2020.113663}.
\bibitem[{Martinez-Gil \& Chaves{-}Gonzalez(2021)}]{key-martinez-kbs}
\bibinfo{author}{Martinez-Gil, J.}, \& \bibinfo{author}{Chaves{-}Gonzalez,
  J.~M.} (\bibinfo{year}{2021}).
\newblock \bibinfo{title}{Semantic similarity controllers: On the trade-off
  between accuracy and interpretability}.
\newblock {\it \bibinfo{journal}{Knowl. Based Syst.}\/},  {\it
  \bibinfo{volume}{234}\/}, \bibinfo{pages}{107609}.
  \DOIprefix\doi{10.1016/j.knosys.2021.107609}.
\bibitem[{Martinez-Gil \& Chaves-Gonzalez(2022)}]{key-martinez-jfis}
\bibinfo{author}{Martinez-Gil, J.}, \& \bibinfo{author}{Chaves-Gonzalez, J.~M.}
  (\bibinfo{year}{2022}).
\newblock \bibinfo{title}{Sustainable semantic similarity assessment}.
\newblock {\it \bibinfo{journal}{Journal of Intelligent \& Fuzzy Systems}\/},
  {\it \bibinfo{volume}{43}\/}, \bibinfo{pages}{6163--6174}.
  \DOIprefix\doi{10.3233/JIFS-220137}.
\bibitem[{Mc~Laughlin(1969)}]{mc1969smog}
\bibinfo{author}{Mc~Laughlin, G.~H.} (\bibinfo{year}{1969}).
\newblock \bibinfo{title}{Smog grading-a new readability formula}.
\newblock {\it \bibinfo{journal}{Journal of reading}\/},  {\it
  \bibinfo{volume}{12}\/}, \bibinfo{pages}{639--646}.
\bibitem[{Meade \& Smith(1991)}]{meade1991readability}
\bibinfo{author}{Meade, C.~D.}, \& \bibinfo{author}{Smith, C.~F.}
  (\bibinfo{year}{1991}).
\newblock \bibinfo{title}{Readability formulas: cautions and criteria}.
\newblock {\it \bibinfo{journal}{Patient education and counseling}\/},  {\it
  \bibinfo{volume}{17}\/}, \bibinfo{pages}{153--158}.
\bibitem[{Mikolov et~al.(2013)Mikolov, Sutskever, Chen, Corrado \&
  Dean}]{key-Mikolov}
\bibinfo{author}{Mikolov, T.}, \bibinfo{author}{Sutskever, I.},
  \bibinfo{author}{Chen, K.}, \bibinfo{author}{Corrado, G.~S.}, \&
  \bibinfo{author}{Dean, J.} (\bibinfo{year}{2013}).
\newblock \bibinfo{title}{Distributed representations of words and phrases and
  their compositionality}.
\newblock In {\it \bibinfo{booktitle}{Advances in Neural Information Processing
  Systems 26: 27th Annual Conference on Neural Information Processing Systems
  2013. Proceedings of a meeting held December 5-8, 2013, Lake Tahoe, Nevada,
  United States.}\/} (pp. \bibinfo{pages}{3111--3119}).
\bibitem[{Miller(1995)}]{miller1995wordnet}
\bibinfo{author}{Miller, G.~A.} (\bibinfo{year}{1995}).
\newblock \bibinfo{title}{Wordnet: a lexical database for english}.
\newblock {\it \bibinfo{journal}{Communications of the ACM}\/},  {\it
  \bibinfo{volume}{38}\/}, \bibinfo{pages}{39--41}.
\bibitem[{Mitchell(2018)}]{mitchell2018web}
\bibinfo{author}{Mitchell, R.} (\bibinfo{year}{2018}).
\newblock {\it \bibinfo{title}{Web scraping with Python: Collecting more data
  from the modern web}\/}.
\newblock \bibinfo{publisher}{" O'Reilly Media, Inc."}.
\bibitem[{Navigli \& Martelli(2019)}]{key-Navigli}
\bibinfo{author}{Navigli, R.}, \& \bibinfo{author}{Martelli, F.}
  (\bibinfo{year}{2019}).
\newblock \bibinfo{title}{An overview of word and sense similarity}.
\newblock {\it \bibinfo{journal}{Nat. Lang. Eng.}\/},  {\it
  \bibinfo{volume}{25}\/}, \bibinfo{pages}{693--714}.
  \DOIprefix\doi{10.1017/S1351324919000305}.
\bibitem[{Pantula \& Kuppusamy(2022)}]{key-Pantula}
\bibinfo{author}{Pantula, M.}, \& \bibinfo{author}{Kuppusamy, K.~S.}
  (\bibinfo{year}{2022}).
\newblock \bibinfo{title}{A machine learning-based model to evaluate
  readability and assess grade level for the web pages}.
\newblock {\it \bibinfo{journal}{Comput. J.}\/},  {\it \bibinfo{volume}{65}\/},
  \bibinfo{pages}{831--842}. \URLprefix
  \url{https://doi.org/10.1093/comjnl/bxaa113}.
  \DOIprefix\doi{10.1093/comjnl/bxaa113}.
\bibitem[{Pedersen et~al.(2004)Pedersen, Patwardhan \&
  Michelizzi}]{key-Pedersen}
\bibinfo{author}{Pedersen, T.}, \bibinfo{author}{Patwardhan, S.}, \&
  \bibinfo{author}{Michelizzi, J.} (\bibinfo{year}{2004}).
\newblock \bibinfo{title}{Wordnet: : Similarity - measuring the relatedness of
  concepts}.
\newblock In \bibinfo{editor}{D.~L. McGuinness}, \&
  \bibinfo{editor}{G.~Ferguson} (Eds.), {\it \bibinfo{booktitle}{Proceedings of
  the Nineteenth National Conference on Artificial Intelligence, Sixteenth
  Conference on Innovative Applications of Artificial Intelligence, July 25-29,
  2004, San Jose, California, {USA}}\/} (pp. \bibinfo{pages}{1024--1025}).
\newblock \bibinfo{publisher}{{AAAI} Press / The {MIT} Press}.
\bibitem[{Qin(2021)}]{key-Qin}
\bibinfo{author}{Qin, Y.} (\bibinfo{year}{2021}).
\newblock \bibinfo{title}{Comparable study on readability of machine generated
  news and human news}.
\newblock In {\it \bibinfo{booktitle}{6th International Conference on Control,
  Robotics and Cybernetics, {CRC} 2021, Shanghai, China, October 9-11, 2021}\/}
  (pp. \bibinfo{pages}{339--343}).
\newblock \bibinfo{publisher}{{IEEE}}.
\newblock \URLprefix \url{https://doi.org/10.1109/CRC52766.2021.9620136}.
  \DOIprefix\doi{10.1109/CRC52766.2021.9620136}.
\bibitem[{Rus et~al.(2013)Rus, Lintean, Banjade, Niraula \&
  Stefanescu}]{key-Rus}
\bibinfo{author}{Rus, V.}, \bibinfo{author}{Lintean, M.~C.},
  \bibinfo{author}{Banjade, R.}, \bibinfo{author}{Niraula, N.~B.}, \&
  \bibinfo{author}{Stefanescu, D.} (\bibinfo{year}{2013}).
\newblock \bibinfo{title}{{SEMILAR:} the semantic similarity toolkit}.
\newblock In {\it \bibinfo{booktitle}{51st Annual Meeting of the Association
  for Computational Linguistics, {ACL} 2013, Proceedings of the Conference
  System Demonstrations, 4-9 August 2013, Sofia, Bulgaria}\/} (pp.
  \bibinfo{pages}{163--168}).
\bibitem[{Senter \& Smith(1967)}]{senter1967automated}
\bibinfo{author}{Senter, R.}, \& \bibinfo{author}{Smith, E.~A.}
  (\bibinfo{year}{1967}).
\newblock {\it \bibinfo{title}{Automated readability index}\/}.
\newblock \bibinfo{type}{Technical Report} Cincinnati Univ OH.
\bibitem[{Skianis et~al.(2020)Skianis, Malliaros, Tziortziotis \&
  Vazirgiannis}]{skianis2020boosting}
\bibinfo{author}{Skianis, K.}, \bibinfo{author}{Malliaros, F.~D.},
  \bibinfo{author}{Tziortziotis, N.}, \& \bibinfo{author}{Vazirgiannis, M.}
  (\bibinfo{year}{2020}).
\newblock \bibinfo{title}{Boosting tricks for word mover’s distance}.
\newblock In {\it \bibinfo{booktitle}{International Conference on Artificial
  Neural Networks}\/} (pp. \bibinfo{pages}{761--772}).
\newblock \bibinfo{organization}{Springer}.
\bibitem[{Todirascu et~al.(2016)Todirascu, Fran{\c{c}}ois, Bernhard, Gala \&
  Ligozat}]{todirascu2016cohesive}
\bibinfo{author}{Todirascu, A.}, \bibinfo{author}{Fran{\c{c}}ois, T.},
  \bibinfo{author}{Bernhard, D.}, \bibinfo{author}{Gala, N.}, \&
  \bibinfo{author}{Ligozat, A.-L.} (\bibinfo{year}{2016}).
\newblock \bibinfo{title}{Are cohesive features relevant for text readability
  evaluation?}
\newblock In {\it \bibinfo{booktitle}{26th International Conference on
  Computational Linguistics (COLING 2016)}\/} (pp. \bibinfo{pages}{987--997}).
\bibitem[{Wilkins(1972)}]{wilkins1972linguistics}
\bibinfo{author}{Wilkins, D.~A.} (\bibinfo{year}{1972}).
\newblock {\it \bibinfo{title}{Linguistics in language teaching}\/} volume
  \bibinfo{volume}{111}.
\newblock \bibinfo{publisher}{Edward Arnold London}.
\bibitem[{Zhang \& Li(2007)}]{key-moead}
\bibinfo{author}{Zhang, Q.}, \& \bibinfo{author}{Li, H.}
  (\bibinfo{year}{2007}).
\newblock \bibinfo{title}{{MOEA/D:} {A} multiobjective evolutionary algorithm
  based on decomposition}.
\newblock {\it \bibinfo{journal}{{IEEE} Trans. Evol. Comput.}\/},  {\it
  \bibinfo{volume}{11}\/}, \bibinfo{pages}{712--731}. \URLprefix
  \url{https://doi.org/10.1109/TEVC.2007.892759}.
  \DOIprefix\doi{10.1109/TEVC.2007.892759}.
\bibitem[{Zhu \& Iglesias(2017)}]{key-sematch}
\bibinfo{author}{Zhu, G.}, \& \bibinfo{author}{Iglesias, C.~A.}
  (\bibinfo{year}{2017}).
\newblock \bibinfo{title}{Computing semantic similarity of concepts in
  knowledge graphs}.
\newblock {\it \bibinfo{journal}{{IEEE} Trans. Knowl. Data Eng.}\/},  {\it
  \bibinfo{volume}{29}\/}, \bibinfo{pages}{72--85}. \URLprefix
  \url{https://doi.org/10.1109/TKDE.2016.2610428}.
  \DOIprefix\doi{10.1109/TKDE.2016.2610428}.

\end{thebibliography}

\end{document}